\pdfoutput=1

\documentclass[11pt]{article}

\usepackage[]{acl}

\usepackage{times}
\usepackage{latexsym}

\usepackage[T1]{fontenc}

\usepackage[utf8]{inputenc}

\usepackage{microtype}

\usepackage{inconsolata}

\usepackage{kotex}
\usepackage{graphicx}
\usepackage{graphics}
\usepackage{xspace}
\usepackage{amsfonts}
\usepackage{amsmath}
\usepackage{algorithm}
\usepackage{algpseudocode}
\usepackage{arydshln}
\usepackage{subcaption}
\usepackage{multirow}
\usepackage{booktabs}
\usepackage{pgfplots}
\usepackage{caption}
\usepackage{verbatim}
\usepackage{tcolorbox}
\usepackage{xcolor}
\usepackage{tcolorbox}    	

\newtcolorbox{boxA}{
    boxrule = 1.5pt,
    colframe = black 
}

\title{Instruction Matters: A Simple yet Effective Task Selection for

Optimized Instruction Tuning of Specific Tasks}

\author{Changho Lee{\textsuperscript{1}\thanks{\;\;\;Equal contribution.}} \quad Janghoon Han{\textsuperscript{1$\ast$}} \quad Seonghyeon Ye{\textsuperscript{2}}\\ {\bf Stanley Jungkyu Choi{\textsuperscript{1}} \quad Honglak Lee{\textsuperscript{1}} \quad Kyunghoon Bae{\textsuperscript{1}}} \\
        {\textsuperscript{1}}LG AI Research\quad{\textsuperscript{2}}KAIST\\ 
        \texttt{\{changho.lee,janghoon.han\}@lgresearch.ai} 
        \\
        }

\begin{document}
\maketitle
\begin{abstract}
Instruction tuning has been proven effective in enhancing zero-shot generalization across various tasks and in improving the performance of specific tasks. For task-specific improvements, strategically selecting and training on related tasks that provide meaningful supervision is crucial, as this approach enhances efficiency and prevents performance degradation from learning irrelevant tasks. In this light, we introduce a simple yet effective task selection method that leverages instruction information alone to identify relevant tasks, optimizing instruction tuning for specific tasks. Our method is significantly more efficient than traditional approaches, which require complex measurements of pairwise transferability between tasks or the creation of data samples for the target task. Additionally, by aligning the model with the unique instructional template style of the meta-dataset, we enhance its ability to granularly discern relevant tasks, leading to improved overall performance. Experimental results demonstrate that training on a small set of tasks, chosen solely based on the instructions, results in substantial improvements in performance on benchmarks such as P3, Big-Bench, NIV2, and Big-Bench Hard. Significantly, these improvements surpass those achieved by prior task selection methods, highlighting the superiority of our approach.\footnote{Code, model checkpoints, and data resources are available at \href{https://github.com/CHLee0801/INSTA}{https://github.com/CHLee0801/INSTA}.}


\end{abstract}

\section{Introduction}
Recently, instruction tuning has gained attention as an innovative approach for improving zero-shot performance \cite{T0,Flan,supernatural,instruct-gpt,alpaca,vicuna}. This method offers the advantage of improving a model's generalization capabilities to unseen tasks by training on diverse tasks accompanied by instructions. The robustness of instruction tuning improves as the diversity of training tasks increases \cite{T0,supernatural,flan_collection,Flan_t5}. 
In this light, recent studies have concentrated on broadening the diversity and increasing the number of tasks within the meta-dataset \cite{meta-dataset} used for instruction tuning \cite{supernatural,Flan_t5,self-instruct, dynosaur, enssemble-instruct}.

Besides applying instruction tuning for general unseen tasks, there is also growing interest in instruction tuning as a methodology to improve the performance of specific unseen tasks \cite{specialist-ie,specialist_wrt,specialist-dst}.
Instruction tuning focuses on specific tasks and trains not all but only informative tasks with instruction format. This strategy is based on the insight that not all tasks are helpful to specific tasks, and some tasks could even lead to performance degradation due to negative transfer during multi-task training \cite{bloomz, camel_go,not_all_task,roe,taskweb,specialist}. However, selecting relevant tasks for training presents a significant challenge. First, manually reviewing through the vast array of datasets for instruction tuning is not feasible. Additionally, discerning the relevance of certain tasks in the training dataset to the target task is often ambiguous.

To address this challenge, studies have been conducted to automatically quantify the relevance between tasks \cite{selector_ori,selector_ye,roe,selector-art,taskweb}. These methods are primarily divided into two categories. The first assesses pairwise task transferability by training models on one task and evaluating their performance on another \cite{pairwise1,pairwise2,not_all_task,taskweb}. The second calculates similarity scores between tasks by comparing small samples from each task's dataset \cite{selector_ori,selector_ye,roe,selector-art}. However, the former approach, which measures pairwise transfer for every task, can be exceedingly time-consuming and compute-intensive. Moreover, the latter approach still necessitates the construction of data for unseen target tasks, which not only introduces additional burdens but also does not fully align with the fundamental goal of instruction tuning, which is to enhance zero-shot capabilities.

In this study, we explore a straightforward yet effective method for choosing relevant tasks for optimized instruction tuning. We focus on the feature of instruction tuning, where the instructions\footnote{Following \citet{T0, Flan, supernatural}, instruction means prompt, template, and task description without instance.} define the characteristics of each task. Building upon this, we introduce an \textbf{Ins}truction-based \textbf{Ta}sk selector (\textsc{InsTa}) that leverages instruction similarity scores to determine task relevance and select informative tasks for training. Through experiments, we discover that this simplified \textit{instruction-only} selection method adequately identifies related tasks and significantly improves performance. Moreover, by further aligning \textsc{InsTa} through training on the instruction style and format of a meta-dataset, it is able to closely understand the instructional nuances, achieving significant performance improvements.

A notable finding from our study is that task selection based exclusively on instructions surpasses previous sample-based methods \cite{selector_ori,selector_ye,roe,selector-art} that employ both instructions and instances. Moreover, instruction information alone shows a high correlation with task similarity as determined by complex pairwise transfer methods \cite{not_all_task} and even achieves slightly better average zero-shot performance. This indicates that the instruction-based approach for selecting related tasks is not only easily applicable but also highly effective.

In summary, our contributions are as follows: 
\begin{itemize} 
    \item We introduce an instruction-based task selection method for optimized instruction tuning. This method efficiently identifies relevant tasks without the extensive computation and data construction as in previous approaches.
    \item By aligning \textsc{InsTa} with the instruction styles and formats of meta-datasets, we significantly improve performance, demonstrating the importance of understanding instructional nuances in instruction-based task selection.
    \item Extensive experiments and comprehensive analyses across various benchmarks validate the superiority of our methodology, showcasing its enhanced efficiency, effectiveness, and practical applicability compared to previous methods.
\end{itemize}


\section{Related Work}
\subsection{Instruction Tuning: Generalist or Specialist}
Instruction tuning shows remarkable zero-shot performance on unseen tasks by training models on various tasks integrated with corresponding instructions. This approach can be broadly categorized into two main streams: instruction tuning as a generalist, which aims to perform well across various unseen tasks \cite{Flan,T0,supernatural,instruct-gpt,alpaca,vicuna}, and instruction tuning as a specialist, which focuses on excelling in specific tasks rather than achieving proficiency across all tasks \cite{specialist-ie,specialist-dst,instruction-retrieval}, as denoted by previous research \cite{specialist}.

According to \citet{do_model}, instruction tuning as a generalist can be categorized based on its objectives: generalizing to "unseen tasks" and generalizing to "unseen instructions". Early studies propose the approach of generalizing to unseen tasks, which involves training on various natural language processing (NLP) tasks and evaluating on unseen tasks \cite{Flan,T0,supernatural}. However, the task-based approach to instruction tuning faces limitations as it is challenging to define all instructions as corresponding tasks, making it difficult to generalize to diverse, user-oriented instructions. In response, a methodology that trains LLMs on diverse instructions without clear task boundaries has been proposed, aiming to generalize to unseen instructions rather than unseen tasks \cite{instruct-gpt,alpaca,vicuna}.

Recent trends have seen the emergence of research focusing on enhancing the zero-shot capabilities of specific tasks through instruction tuning \cite{specialist_wrt,specialist-ie,specialist-dst}. Instead of learning all tasks, the model selectively learns tasks related to the target task in instruction format, aiming to perform well on specific target tasks. Notably, \citet{specialist} demonstrate that training exclusively on tasks related to a specific target task outperforms the instruction tuning as a generalist in terms of performance. 

\subsection{Quantifying Task Relationship }
Research on understanding which tasks can be helpful for other tasks has been extensively conducted across various fields. \citet{vu_transfer1,poth_transfer,yada_transfer} calculate intermediate task transfer scores to discern the relationships between tasks, aiming to determine beneficial tasks to train the model before fine-tuning on specific target tasks. Additionally, \citet{vu_transfer2,su_transfer} measure tasks that are helpful in parameter-efficient tuning through prompt transfer.

The research on identifying task relationships has been extended to the field of instruction tuning as well. \citet{not_all_task,taskweb} measure pairwise task transfer between every task pair to identify helpful source tasks on specific target tasks. On the other hand, \citet{roe,selector-art} have calculated task similarity using only a few data samples from training and evaluation tasks instead of training and evaluating every task pair individually.

Previous task selection approaches, which automatically measure task similarities, are time-consuming because they require training and evaluating every task pair or still rely on the availability of data samples. Our method simplifies this process by evaluating task relationships solely through instructions, eliminating the need for laborious measurements across every task or generating test data. For more detailed descriptions of our method's practicalities and efficiency compared to prior works, please refer to Appendices \ref{sec:previous_method} and \ref{sec:efficiency}.

\section{Instruction-based Task Selector (\textsc{InsTa})}
To enhance the zero-shot capability of LLMs for specific target tasks, we select informative tasks that positively impact the performance of the target task. Our task selection method exclusively relies on instruction information to assess the relevance between tasks, which is efficient as it removes the necessity of measuring correlations between all training and evaluation task pairs. 
Furthermore, unlike previous methods, our approach has the advantage of easy applicability with just the instruction (task description) of the target task, without the need for constructing data samples for the target task.

\subsection{Formulation}
The meta-dataset $\mathbf{M}$ consists of multiple task clusters $\mathbf{C}$ and each task cluster comprises several tasks $\mathbf{T}$ with the same task type. Each task $\mathbf{T}$ includes various instructions $\mathbf{I}$ and corresponding instances. We aim to find tasks related to the target task $\mathbf{\bar{T}}$. To identify tasks related to the target task $\mathbf{\bar{T}}$, we measure instruction-based task similarity score as follows:
\begin{equation}
\mathbf{Score}(I^{\bar{T}}_i,I_j^{T}) = \cos(\mathit{E}(I^{\bar{T}}_i),\mathit{E}(I_j^{T}))
\label{eq:score}
\end{equation}
where \( I^{\bar{T}}_i \) denotes the \( i^{th} \) instruction of the target task \( \bar{T} \), and \( I_j^{T} \) denotes the \( j^{th} \) instruction of some arbitrary task $T \in M$ chosen for similarity assessment. For measuring similarity, we employ cosine similarity, and for the embedding function \( E \), we utilize the Sentence Transformer \cite{sentence-bert} as an off-the-shelf embedding model, following \citet{roe}. For more specific details, please refer to the Appendix \ref{off-the-shelf embedding model}.

\subsection {Aligning Instruction-based Task Selector with Meta-dataset}
\label{task_selector_explanation}
 The off-the-shelf embedding model often lacks the capability to accurately identify related tasks based on instructions, as it is not trained in the unique instruction styles present in meta-datasets. To mitigate this issue, our approach includes an additional aligning process that fine-tunes our selector model to adapt to the distinctive instruction styles of each meta-dataset. For training, we select a random instruction from the same task as the given instruction as a positive sample and designate instructions from different task clusters as negative samples.\footnote{The rationale behind not utilizing task instructions from the same task cluster as positive or negative samples is twofold: in the former case, even though tasks may have the same task type, they can still be clearly differentiated, and in the latter case, false negatives may occur, even within the same task type.} The training objective is as follows:
\begin{equation}
\label{eq:instruction_loss}
L(I_i, I_j, y) = \left( y - \mathbf{Score}(I_i, I_j) \right)^2
\end{equation}
where $y \in \{0,1\}$ denotes the truth label of a given instruction pair $(I_i, I_j)$. $y$ = 1 indicates that $I_j$ is in same task; otherwise, $y$ = 0. The similarity score is measured using Equation \ref{eq:score}.

\subsection{Multi-task Selection for Instruction Tuning}
To efficiently perform instruction tuning for a specific target task, we select the most relevant training tasks by employing an \textsc{InsTa} model as detailed in Section \ref{model_explanation}.
In this process, for each instruction \( I_i^{\bar{T}} \) of the unseen target task \(\bar{T}\), we compute the similarity scores with every instruction \( I_j^{T} \) across all tasks \( T \) in the training set, as defined by Equation \ref{eq:score}. Based on these computed scores, we then select the top-$k$ tasks \( T \) that exhibit the highest degrees of similarity. This selection mechanism is encapsulated by the following formula:
\begin{equation}
k\text{-argmax}_{T} \left\{ \mathbf{Score}(I_i^{\bar{T}}, I_j^{T}) \,, \forall{i, j} \right\}
\label{eq:task_selection}
\end{equation}
where the "$k$-argmax" operation in the formula signifies selecting the top-$k$ tasks that have the highest scores.

\section{Experimental Setup}
\subsection{Dataset}
We conduct experiments on two representative instruction tuning meta-datasets: P3 \textit{(Public Pool of Prompts)} \cite{T0,promptsource} and NIV2 \textit{(SuperNaturalInstructions V2)} \cite{supernatural}. P3 is a meta-dataset comprised of 12 task clusters. It contains a total of 35 tasks across 8 task clusters for training and 11 tasks across 4 task clusters for held-out evaluation. Note that each task in P3 includes 11.7 instructions on average. Conversely, NIV2 encompasses 72 task clusters of English tasks. The training set comprises a total of 756 tasks across 60 clusters, while the held-out evaluation tasks include 119 tasks across 12 task clusters. In the case of NIV2, each task consists of a single instruction. We additionally evaluate BIG-Bench \cite{bigbench} and BIG-Bench Hard (BBH) \cite{bigbench_hard} as supplementary evaluation datasets. For more detailed information, please refer to Table \ref{table:statistics} and Appendix \ref{sec:examples_of_p3_niv2_instruction}.

\begin{table}[]
\centering
    \resizebox{\columnwidth}{!}{
\begin{tabular}{lcc}
    \toprule
       Statistics & \textsc{P3} & \textsc{NIV2} \\
    \midrule
        \# of training tasks & 35 & 756 \\
        \# of training task clusters & 8 & 63 \\
        Avg. \# of instructions (per task) & 8.45 & 1(+1) \\
        Max \# of training instance (per task) & 50,000 & 5,000 \\
        \# of selected task for training & 5 & 70 \\
    \midrule
        \# of evaluation tasks & 11 & 33 \\
        \# of evaluation task clusters & 4 & 12 \\
        Additional evaluation & Big-Bench & BBH \\
        \# of additional evaluation tasks & 14 & 27 \\
        Evaluation metric & ACC & ROUGE-L \\
    \bottomrule
\end{tabular}}
\caption{Statistics of P3 and NIV2. (+1) in Avg. \# of instructions for NIV2 represents GPT-4 augmented instruction.}
\label{table:statistics}
\end{table} 

\subsection{Task Selector Setup}
\label{task_selection_setup}

\begin{table*}[h!]
    \resizebox{\textwidth}{!}{\begin{tabular}{lcccccccccccc}
    \toprule
    \multirow{2}{*}{\textbf{Method}} & \multicolumn{5}{c}{NLI} & \multicolumn{3}{c}{Sentence Completion} & \multicolumn{2}{c}{Coref. Resol.} & WSD & \multirow{2}{*}{\textbf{Total Avg.}}
    \\ \cmidrule(lr){2-6} \cmidrule(lr){7-9} \cmidrule(lr){10-11} \cmidrule(lr){12-12} & \textbf{RTE} & \textbf{CB} & \textbf{AN. R1} & \textbf{AN. R2} & \textbf{AN. R3} & \textbf{COPA} & \textbf{Hellasw.} & \textbf{StoryC.} & \textbf{Winogr.} & \textbf{WSC} & \textbf{WiC} &  \\
    \midrule
    T0-11B & 80.83 & 70.12 & 43.56 & 38.68 & 41.26 & 90.02 & 33.58 & 92.40 & 59.94 & 61.45 & 56.58 & 60.77 \\
    GPT-3(175B) & 63.50 & 46.40 & 34.60 & 35.40 & 34.50 & 91.00 & 78.90 & 83.20 & 70.20 & 65.40 & 45.92 & 59.00 \\
    \midrule
    T5(3B) & 54.37 & 36.73 & 33.13 & 33.67 & 32.83 & 60.13 & 23.35 & 46.30 & 50.29 & 41.35 & 50.80 & 41.61 \\
    T0-3B & 60.61 & 44.64 & 35.17 & 33.37 & 33.55 & 74.75 & 27.42 & 84.82 & 50.84 & 63.22 & 51.21 & 50.87 \\
    T5(3B) + Random & 53.07 & 44.13 & 33.13 & 33.61 & 34.02 & 62.38 & 27.92 & 51.48 & 51.66 & 41.35 & 50.58 & 43.94 \\
    T5(3B) + Pairwise Transfer & 64.95 & \textbf{58.42} & \textbf{39.17} & \underline{35.90} & \textbf{41.73} & 90.13 & \underline{30.59} & \textbf{97.37} & \underline{60.19} & \underline{63.70} & \textbf{54.33} & \underline{57.86} \\
    T5(3B) + \textsc{PE w/ RoE} & 64.01 & 43.57 & 35.49 & 34.64 & 31.22 & 79.25 & 34.60* & 86.33 & \textbf{61.60} & 62.21 & \underline{52.97} & 53.26 \\
    T5(3B) + \textsc{InsTa} & \underline{73.86} & 55.10 & 36.82 & 34.77 & 35.27 & \underline{91.00} & 27.63 & 94.10 & 55.26 & 56.13 & 52.84 & 55.70 \\
    T5(3B) + \textsc{InsTa}$_{\textbf{\textit{Aligned}-P3}}$ & \textbf{77.87} & \underline{56.89} & \underline{38.28} & \textbf{36.30} & \underline{37.18} & \textbf{92.50} & \textbf{31.40} & 95.86 & 56.37 & \textbf{64.42} & 50.61 & \textbf{57.97} \\
    \bottomrule
    \end{tabular}}
\caption{Evaluation performance on P3 datasets. We report the performance of 11 different unseen datasets categorized into 4 task categories. We select top-5 datasets from pairwise transfer results from \citet{not_all_task} for T5(3B) + Pairwise Transfer model, which measured transferability from every source task to every target task. \textsc{PE w/ RoE} represents Prompt Experts with Retrieval of Experts (RoE) from \cite{roe}. Note that Hellaswag* performance from \citet{roe} includes auxiliary tasks, showing comparably higher performance. The best comparable performances are \textbf{bolded} and second best \underline{underlined}.}
\label{table:p3_selector}
\vspace{-3mm}
\end{table*}

\paragraph{P3} \label{p3_refinement} consists of an average of 11.7 instructions per task. In P3, there are some instructions that are designed to diversify task types in a single dataset\footnote{For example, "generating documents from its summary" for summarization task.}, and we exclude such instructions from selector training since they may hinder the selection of relevant tasks. Furthermore, instructions for P3 tasks include unique placeholders\footnote{\{\{-\}\} represents placeholder in the instruction. An example instruction from the WiC task: "Does the word \{\{word\}\} have the same meaning in these two sentences? Yes, No?\textbackslash n\{\{sentence1\}\}\textbackslash n\{\{sentence2\}\}.\label{wic_instruction}}. These placeholders could act as misleading shortcuts and may negatively influence training. Therefore, we have standardized them to \{\{text\}\} and \{\{candidate\}\} for input snippets and label space, respectively. Please refer to the Appendix \ref{sec:p3_instruction_formulation} for more detail about P3 instruction formulation.
\paragraph{NIV2} \label{niv2_setup} instructions feature human-crafted, human-readable \textit{Task Definition}, \textit{Positive Task Examples}, \textit{Negative Task Examples}, and \textit{Explanation}. We utilize the \textit{Task Definition} for training the task selector. Moreover, since NIV2 has only one instruction per task, there are no positive samples for training the task selector. To address this, we generate instructions for all NIV2 tasks using GPT-4 \cite{gpt4}. Specifically, we generate a paraphrased one and employ it as a positive sample for training the task selector. For more information about a query used for GPT-4 API calls and generated instructions, please refer to the Appendix \ref{sec:niv2_instruction_generation}.

\begin{table}[ht!]
    \resizebox{\columnwidth}{!}
    {\begin{tabular}{lccc|ccc}
    \toprule
    \multirow{3}{*}{Dataset (metric)} & \multirow{2}{*}{T0} & \multirow{2}{*}{Cos PE} & T5(3B) + & \multirow{2}{*}{T0} & \multirow{2}{*}{GPT-3} & \multirow{2}{*}{PaLM} \\
    & & & \textsc{InsTa} & & & \\
    & 3B & 3B & $_{\textbf{\textit{Aligned}-P3}}$ & 11B & 175B & 540B \\ 
    \midrule
    Known Un. & 50.00 & \underline{58.70} & \textbf{65.22} & 65.22 & 60.87 & 56.52 \\
    Logic Grid & \underline{32.90} & 30.70 & \textbf{35.40} & 33.67 & 31.20 & 32.10 \\
    Strategy. & \underline{53.06} & 42.36 & \textbf{55.50} & 54.67 & 52.30 & 64.00 \\
    Hindu Kn. & 35.43 & \underline{51.43} & \textbf{58.29} & 42.86 & 32.57 & 56.00 \\
    Movie D. & \textbf{52.84} & 46.72 & \underline{52.48} & 57.33 & 51.40 & 49.10 \\
    Code D. & 43.33 & \textbf{66.67} & \underline{51.67} & 51.67 & 31.67 & 25.00 \\
    Concept & 63.18 & \underline{72.82} & \textbf{78.34} & 71.72 & 26.78 & 59.26 \\
    Language & 15.08 & \textbf{25.95} & \underline{21.31} & 18.33 & 15.90 & 20.10 \\
    Vitamin & \underline{61.28} & 46.55 & \textbf{64.77} & 57.33 & 12.30 & 14.10 \\
    Syllogism & \textbf{51.08} & 50.00 & 50.94 & 48.33 & 50.50 & 49.90 \\
    Misconcept. & \underline{52.05} & 47.03 & \textbf{53.42} & 52.97 & 47.95 & 47.47 \\
    Logical & \underline{43.18} & 42.40 & \textbf{46.06} & 54.67 & 23.42 & 24.22 \\
    Winowhy & \underline{44.29} & \textbf{44.33} & \textbf{44.33} & 55.00 & 51.50 & 45.30 \\
    Novel Con. & \underline{21.88} & - & \textbf{25.00} & 28.13 & 46.88 & 46.88 \\
    \midrule
    BIG-bench AVG & 44.26 & \underline{48.13*} & \textbf{50.20} & 51.06 & 37.57 & 41.77 \\
    \bottomrule
    \end{tabular}}
\caption{Evaluation performance on 13 BIG-bench tasks. \textsc{Cos PE} represents the \textsc{PE} trained on \textsc{Cosmos-qa} from \citet{roe}. Note that the average performance of \textsc{Cos PE} excludes the score for \textsc{Novel Concepts} since it is not publicly available. The best comparable performances are \textbf{bolded} and second best \underline{underlined}.}
\label{table:p3_bigbench}
\vspace{-3mm}
\end{table}

\subsection{Training Setup}
Following previous studies \cite{T0,supernatural}, we employ all tasks, excluding those held out for evaluation, as our training set. In contrast to traditional approaches that train on all tasks, our strategy specifically trains only the top-$k$ tasks considered the most informative for each target task.

For the P3, we select the top-5 highest scoring training tasks out of 35 tasks for each target task. For each training, we randomly sample 50k instances for each training task, totaling 250k instances. Unlike the P3, each task in NIV2 has only one instruction, and data instances are limited to 6.5k. Considering this difference, we select top-70 tasks out of 756 tasks for each target task and randomly sample up to 5k instances for each task, which ends up with 350k instances in total. These training quantities are significantly smaller than the 2M instances of T0 \cite{T0} and 5.7M instances of \textsc{T\textit{k}-Instruct} \cite{supernatural}, which correspond to the instruction-tuned models on P3 and NIV2, respectively. 

We use the T5 LM-adapted model(3B) \cite{t5-lm-adapted} as our base model and train for 3 epochs with a constant learning rate of 1e-4 and 5e-5 for P3 and NIV2, respectively. We configure validation from training datasets and select the model that shows the best performance in the validation. Our experiment is in a true zero-shot setting, where no samples from the held-out task are used for checkpoint selection. Please refer to the Appendix \ref{training_setup} for more detail.

\begin{figure*}[ht!]
\centering
    \includegraphics[width=0.92\textwidth]{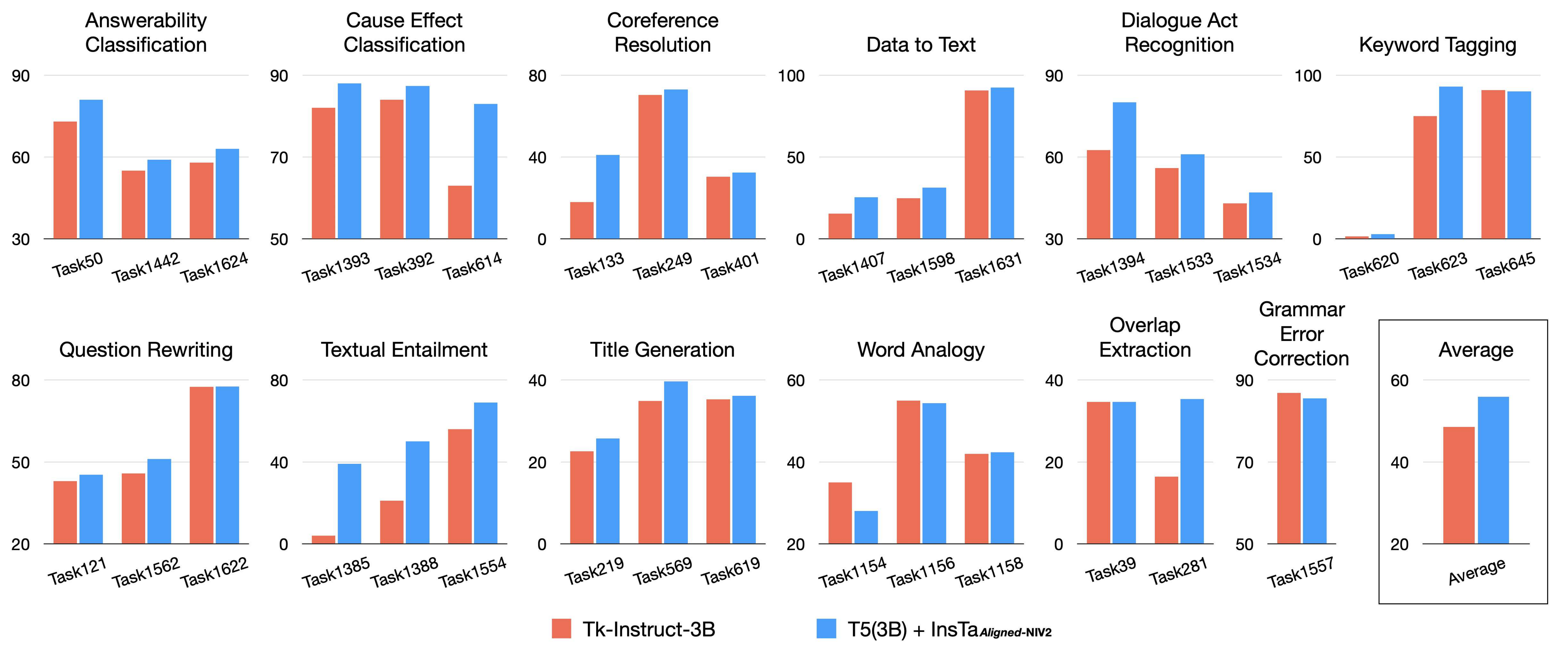}
\caption{Evaluation performance on NIV2 datasets. We report the performance of 33 different tasks from 12 different task clusters, and the average performance of \textsc{T\textit{k}-Instruct-3B} and T5(3B) + \textsc{InsTa}$_{\textbf{\textit{Aligned}-NIV2}}$. We use a \textsc{T\textit{k}-Instruct} model trained on [\textit{Def + Pos(2)}] setting.}
\label{fig:supernat_selector}
\vspace{-3mm}
\end{figure*}  

\paragraph{Models}
\label{model_explanation}
For the P3 dataset, we use the following baselines. \textbf{GPT-3(175B)} is an autoregressive LM that has shown remarkable ability in following demonstrations provided in its instructions \cite{gpt3}. \textbf{T0-11B / T0-3B} have same architecture as T5 but trained on millions of samples from 35 different P3 tasks \cite{T0}. \textbf{T5(3B)} is text-to-text pretrained LM without instruction tuning \cite{T5}. \textbf{T5(3B) + Random} is trained on 5 random tasks from P3. \textbf{T5(3B) + Pairwise Transfer} is trained on top-5 tasks with the highest transferability scores demonstrated in \citet{not_all_task}, representing pairwise transfer task selection. \textbf{T5(3B) + PE \textsc{w/RoE}} denotes a model that selects relevant tasks by measuring cosine similarity between training and evaluation data samples \cite{roe}. 

For NIV2, we set  \textbf{\textsc{T\textit{k}-Instruct-3B}} as our baseline, which has the same architecture as T5-3B but trained on millions of samples from 756 different NIV2 tasks. 

Finally, to demonstrate the effectiveness of instruction-based task selection, we introduce the following model variants:
\begin{itemize}
    \item \textbf{T5(3B) + \textsc{InsTa}}: Model trained on selected top-$k$ tasks by \textbf{\textsc{Ins}}truction-based \textbf{\textsc{Ta}}sk Selector (Off-the-shelf embedding model).
    \item \textbf{T5(3B) + \textsc{InsTa}$_{\textbf{\textit{Aligned}-P3/NIV2}}$}: Models trained on selected top-$k$ tasks by further aligned \textbf{\textsc{InsTa}} on instructions from P3 and NIV2, respectively.
\end{itemize}

\begin{table}[t!]
    \resizebox{\columnwidth}{!}{\begin{tabular}{lcc}
    \toprule
    \multirow{2}{*}{Dataset (metric)} & \multirow{2}{*}{\textsc{T\textit{k}-Instruct-3B}} & T5(3B) + \\
    & & \textsc{InsTa}$_{\textbf{\textit{Aligned}-NIV2}}$ \\
    \midrule
    Boolean Expressions & 54.00 & \textbf{63.20} \\
    Causal Judgement & \textbf{58.29} & \textbf{58.29} \\ 
    Date Understanding & 26.00 & \textbf{33.20} \\
    Disambiguation QA & 44.00 & \textbf{53.80} \\
    Dyck Languages & \textbf{0.00} & \textbf{0.00} \\
    Formal Fallacies & 53.60 & \textbf{53.80} \\
    Geometric Shapes & 0.24 & \textbf{9.20} \\
    Hyperbaton & 48.40 & \textbf{49.60} \\
    Logical Deduction Three Objects & 13.51 & \textbf{47.20} \\
    Logical Deduction Five Objects & 17.71 & \textbf{31.6} \\
    Logical Deduction Seven Objects & 16.23 & \textbf{30.00} \\
    Movie Recommendation & 23.20 & \textbf{60.40} \\
    Multistep Arithmetic & 0.55 & \textbf{3.60} \\
    Navigate & \textbf{58.00} & 53.60 \\
    Object Counting & 35.60 & \textbf{38.40} \\
    Penguins in a Table & \textbf{28.76} & 24.66 \\
    Reasoning about Colored Objects & 24.80 & \textbf{29.40} \\
    Ruin Names & \textbf{29.60} & 28.40 \\
    Salient Translation Error Detection & 14.00 & \textbf{28.80} \\
    Snarks & 46.63 & \textbf{57.30} \\
    Sports Understanding & \textbf{52.80} & \textbf{52.80} \\
    Temporal Sequences & 18.80 & \textbf{20.80} \\
    Tracking Shuffled Objects (3) & 32.40 & \textbf{33.20} \\
    Tracking Shuffled Objects (5) & \textbf{17.20} & 14.80 \\
    Tracking Shuffled Objects (7) & \textbf{13.20} & 12.00 \\
    Web of Lies & 51.60 & \textbf{55.60} \\
    Word Sorting & 44.79 & \textbf{44.88} \\
    \midrule
    BBH Average & 30.52 & \textbf{36.61} \\
    \bottomrule
    \end{tabular}}
\caption{Evaluation performance on BigBench-Hard(BBH) tasks. We generate instruction (\textit{Task Definition}) of each task using GPT-4 API. The query used for GPT-4 API and all the generated instructions are presented in the Appendix \ref{sec:bbh_generation} The best comparable performances are \textbf{bolded}.}
\label{table:bbh}
\vspace{-3mm}
\end{table}

\begin{table*}[h!]
    \centering
    \resizebox{0.98\textwidth}{!}{\begin{tabular}{lcccccccccccc}
    \toprule
    \multirow{2}{*}{\textbf{Method}} & \multicolumn{5}{c}{NLI} & \multicolumn{3}{c}{Sentence Completion} & \multicolumn{2}{c}{Coref. Resol.} & WSD & \multirow{2}{*}{\textbf{Total Avg.}}
    \\ \cmidrule(lr){2-6} \cmidrule(lr){7-9} \cmidrule(lr){10-11} \cmidrule(lr){12-12} & \textbf{RTE} & \textbf{CB} & \textbf{AN. R1} & \textbf{AN. R2} & \textbf{AN. R3} & \textbf{COPA} & \textbf{Hellasw.} & \textbf{StoryC.} & \textbf{Winogr.} & \textbf{WSC} & \textbf{WiC} &  \\
    \midrule
    T0-3B & \underline{60.61} & 44.64 & 35.17 & 33.37 & 33.55 & \underline{74.75} & 27.42 & 84.82 & 50.84 & \textbf{63.22} & 51.21 & \underline{50.87} \\
    T5(3B) + \textsc{DSTa} & 54.44 & \textbf{55.36} & \underline{35.41} & \underline{33.99} & \underline{34.14} & 73.63 & \textbf{29.97} & \textbf{94.27} & \underline{54.68} & 39.54 & \underline{51.36} & 50.62 \\
    T5(3B) + \textsc{InsTa} & \textbf{73.86} & \underline{55.10} & \textbf{36.82} & \textbf{34.77} & \textbf{35.27} & \textbf{91.00} & \underline{27.63} & \underline{94.10} & \textbf{55.26} & \underline{56.13} & \textbf{52.84} & \textbf{55.70} \\
    \bottomrule
    \end{tabular}}
\caption{Comparison between \textbf{D}ata \textbf{S}ample-based \textbf{Ta}sk Selector(\textsc{DSTa}) and \textbf{Ins}truction-based \textbf{Ta}sk Selector (\textsc{InsTa}). The best comparable performances are \textbf{bolded} and second best \underline{underlined}.}
\label{table:instance_instruction}
\vspace{-3mm}
\end{table*}

\subsection{Evaluation Setup}
For P3 evaluation, following the evaluation method from \citet{T0}, we apply rank classification and measure the model's performance on every instruction of the target task. We then calculate the average performance for the task. Note that each target task in P3 has 10.09 instructions on average. For NIV2 evaluation, we follow the same evaluation protocol as in \citet{supernatural} and report ROUGE-L \cite{rouge} score. We adopt greedy decoding with a maximum generation length of 256. 

\section{Result}
\label{sec:result}
\subsection{P3 Results}
\label{sec:P3 Results}
Table \ref{table:p3_selector} shows experimental results on the 11 unseen datasets from P3. Compared to T0-3B, which is fully instruction-tuned on all 35 tasks from P3, the T5(3B) + Random model shows inferior performance, but both T5(3B) + \textsc{InsTa} model and T5(3B) + \textsc{InsTa}$_{\textbf{\textit{Aligned}-P3}}$ model outperform 10 tasks out of 11 tasks, with each exhibiting a performance gap of 4.83\% and 7.10\% on average, respectively. They also show superior performance compared to T5(3B) + \textsc{PE w/ RoE}, highlighting the instruction itself is sufficient enough to choose informative tasks without the use of data samples. Moreover, T5(3B) + \textsc{InsTa}$_{\textbf{\textit{Aligned}-P3}}$ outperforms T5(3B) + \textsc{InsTa} across 10 tasks, suggesting further aligning task selector on instructions from the meta-dataset increases the precision of task selection. Additionally, our approach exhibits marginally superior performance compared to T5(3B) + Pairwise Transfer, suggesting that instruction-only task selection can effectively identify related tasks without the need for exhaustive pairwise evaluations. Notably, we observe a strong correlation between pairwise task transferability and our task selection approach, further elaborated in Appendix \ref{pairwise and i-bts selected choice}.


\subsection{Big-Bench Results}
\label{sec:BB Results}
Table \ref{table:p3_bigbench} evaluates the performance of instruction-based task selection on the 14 Big-Bench tasks employing P3 held-in tasks in Section \ref{sec:P3 Results} as the training dataset. We compare our method to baselines T0-3B and T5(3B) + \textsc{Cos PE} of \citet{roe}. Our model, T5(3B) + \textsc{InsTa}$_{\textbf{\textit{Aligned}-P3}}$, surpasses T0-3B in 12 out of 14 tasks, achieving an average performance increase of 5.94\%. Tasks like Movie D. and Syllogism, where results are comparable to random guesses, indicate informative tasks are insufficient in P3. When compared to \textsc{Cos PE} \cite{roe}, our T5(3B) + \textsc{InsTa}$_{\textbf{\textit{Aligned}-P3}}$ model shows improvements in most tasks, with an average increase of 2.07\%. Notably, it also surpasses T0-11B in a majority of tasks. These findings demonstrate that instruction-based task selection enables more effective training even with a small number of tasks.

\subsection{NIV2 Results}
\label{sec:NIV2 Results}
Figure \ref{fig:supernat_selector} illustrates the experimental results for 33 tasks within NIV2\footnote{Though NIV2 has 12 task clusters and 119 tasks for evaluation, we randomly select up to 3 tasks per task cluster due to computational cost.}. We adhere to [\textit{Def + Pos(2)}] setting in NIV2, which includes \textit{Task Definition} and two \textit{Positive Task Examples} in the instruction, but note that we only use \textit{Task Definition} for task selection. The precise input format utilized for training and evaluation is elaborated in the Appendix \ref{niv2_input_formats}.

The experimental result reveals our T5(3B) + \textsc{InsTa}$_{\textbf{\textit{Aligned}-NIV2}}$ surpasses the baseline in most tasks, highlighting the efficacy of our method in NIV2 meta-dataset. We observe notable improvements in performance for specific tasks, such as \textsc{Task614}, \textsc{Task1385}, and \textsc{Task1388}. This suggests that training all tasks could lead to performance degradation in certain tasks. However, training informative tasks selected by our method can alleviate this problem.

\subsection{Big-Bench Hard Results}
\label{sec:BBH Results}
To further validate the effectiveness of our approach, we assess our T5(3B) + \textsc{InsTa}$_{\textbf{\textit{Aligned}-NIV2}}$ model on 27 BBH tasks. This model is trained on the top-70 tasks chosen from 756 NIV2 tasks. In contrast to Section \ref{sec:NIV2 Results}, 
we adopt the \textsc{T\textit{k}-Instruct-3B} [\textit{Def}] setting as our baseline due to the lack of \textit{Positive Task Examples} in the BBH datasets. Moreover, given the absence of \textit{Task Definition} for the BBH tasks, we utilize the GPT-4 API to generate definition for each task.

Consistent with results highlighted in Section \ref{sec:NIV2 Results}, we observe performance enhancements across most tasks. Tasks such as Movie Recommendation, which exhibit suboptimal performance with \textsc{T\textit{k}-Instruct-3B}, demonstrate enhanced performance when only informative tasks are learned. This outcome underscores the robustness of T5(3B) + \textsc{InsTa}$_{\textbf{\textit{Aligned}-NIV2}}$ in mitigating negative transfer by exclusively learning relevant tasks.

\section{Further Analysis}
For further analysis, we conduct experiments to compare instruction-based versus data sample-based task selection, examine the effects of instruction refinement in the P3 dataset, and assess the impacts of varying the number of related tasks selected ($k$). Additionally, changes in task selection performance based on the number of instructions used for alignment, as well as experiments on learning different styles of meta-datasets and their impact on performance, are detailed in Appendices \ref{sec:scale_instruction} and \ref{sec:diff_meta_dataset}.

\subsection{Instruction-based vs. Data Sample-based Task Selection}
\label{sec:temp-inst}
We conduct experiments shown in Table \ref{table:instance_instruction} to compare our instruction-based approach with the traditional method \cite{selector_ye,roe} of selecting related tasks using data samples (integration of instruction and instance). The T5(3B) + \textsc{DSTa} model identifies relevant tasks similar to T5(3B) + \textsc{InsTa}, with the key distinction being the use of data samples for similarity comparison.

While the T5(3B) + \textsc{DSTa} model outperforms on specific tasks like Hellaswag, it falls short in others, including RTE and WSC. Conversely, T5(3B) + \textsc{InsTa} consistently enhances performance across the majority of tasks relative to T0-3B, and achieves an impressive average performance enhancement of 5.08\% over T5(3B) + \textsc{DSTa}. This outcome suggests that instances within data samples might obstruct the extraction of representative task features, diminishing the task selector's ability to identify related tasks effectively. Please refer to Appendix \ref{sec:data_sample_based_experimental_detail} for more experimental details.

\begin{table}[t!]
\centering
    \resizebox{0.95\columnwidth}{!}{\begin{tabular}{lccccc}
    \toprule
    \multirow{2}{*}{\textbf{Task}} & \multirow{2}{*}{T0-3B} & \multicolumn{2}{c}{T5(3B) + \textsc{InsTa}} & \multicolumn{2}{c}{T5(3B) + \textsc{InsTa}$_{\textbf{\textit{Aligned}-P3}}$} \\
    \cmidrule(lr){3-4} \cmidrule(lr){5-6} & & Unfiltered & Filtered & Unfiltered & Filtered \\
    \midrule
    RTE & 60.61 & 73.39 & \textbf{73.86} & 74.69 & \textbf{77.87} \\
    CB & 44.64 & 50.26 & \textbf{55.10} & 54.08 & \textbf{56.89} \\
    Anli R1 & 35.17 & 34.00 & \textbf{36.82} & 35.08 & \textbf{38.28} \\
    Anli R2 & 33.37 & 33.91 & \textbf{34.77} & 34.72 & \textbf{36.3} \\
    Anli R3 & 33.55 & 33.97 & \textbf{35.27} & 35.19 & \textbf{37.18} \\
    COPA & 74.75 & 82.63 & \textbf{91.00} & 87.50 & \textbf{92.5} \\
    Hellaswag & 27.42 & \textbf{36.31} & 27.63 & \textbf{36.46} & 31.4 \\
    StoryCloze & 84.82 & 43.27 & \textbf{94.10} & 48.42 & \textbf{95.86} \\
    Winogrande & 50.84 & \textbf{55.54} & 55.26 & 55.01 & \textbf{56.37} \\
    WSC & 63.22 & \textbf{64.06} & 56.13 & 64.06 & \textbf{64.42} \\
    WiC & 51.21 & 50.69 & \textbf{52.84} & \textbf{51.22} & 50.61 \\
    \midrule
    Average & 50.87 & 50.73 & \textbf{55.70} & 52.4 & \textbf{57.97} \\
    \bottomrule
    \end{tabular}}
\caption{Evaluation performance of P3 datasets before and after the instruction refinement. The best comparable performances are \textbf{bolded}.}
\label{table:filter_unfilter}
\vspace{-3mm}
\end{table}

\subsection{Impact of Instruction Refinement}
\label{sec:temp-refine}
As mentioned in Section \ref{p3_refinement}, our approach operates based on the premise that the instruction accurately describes the characteristics of the task, necessitating the process of instruction refinement. Table \ref{table:filter_unfilter} demonstrates the impact of such instruction refinement on performance. The term "Unfiltered" denotes the training conducted without instruction refinement, while "Filtered" indicates the use of refined models.

Unfiltered models, while showing performance improvements in most tasks compared to T0-3B, encounter performance degradation in certain tasks, notably StoryCloze, due to the selection of irrelevant tasks. Conversely, models that use filtered and refined instructions accurately select related tasks and generally demonstrate improved performance over unfiltered models. This result emphasizes the significance of instruction quality in task selection. Furthermore, the notable performance enhancement seen in T5(3B) + \textsc{InsTa}$_{\textbf{\textit{Aligned}-P3}}$ underscores the substantial role of instruction quality, particularly when additional training for the alignment is applied.

\begin{figure}[t!]
\centering
    \includegraphics[width=\columnwidth]{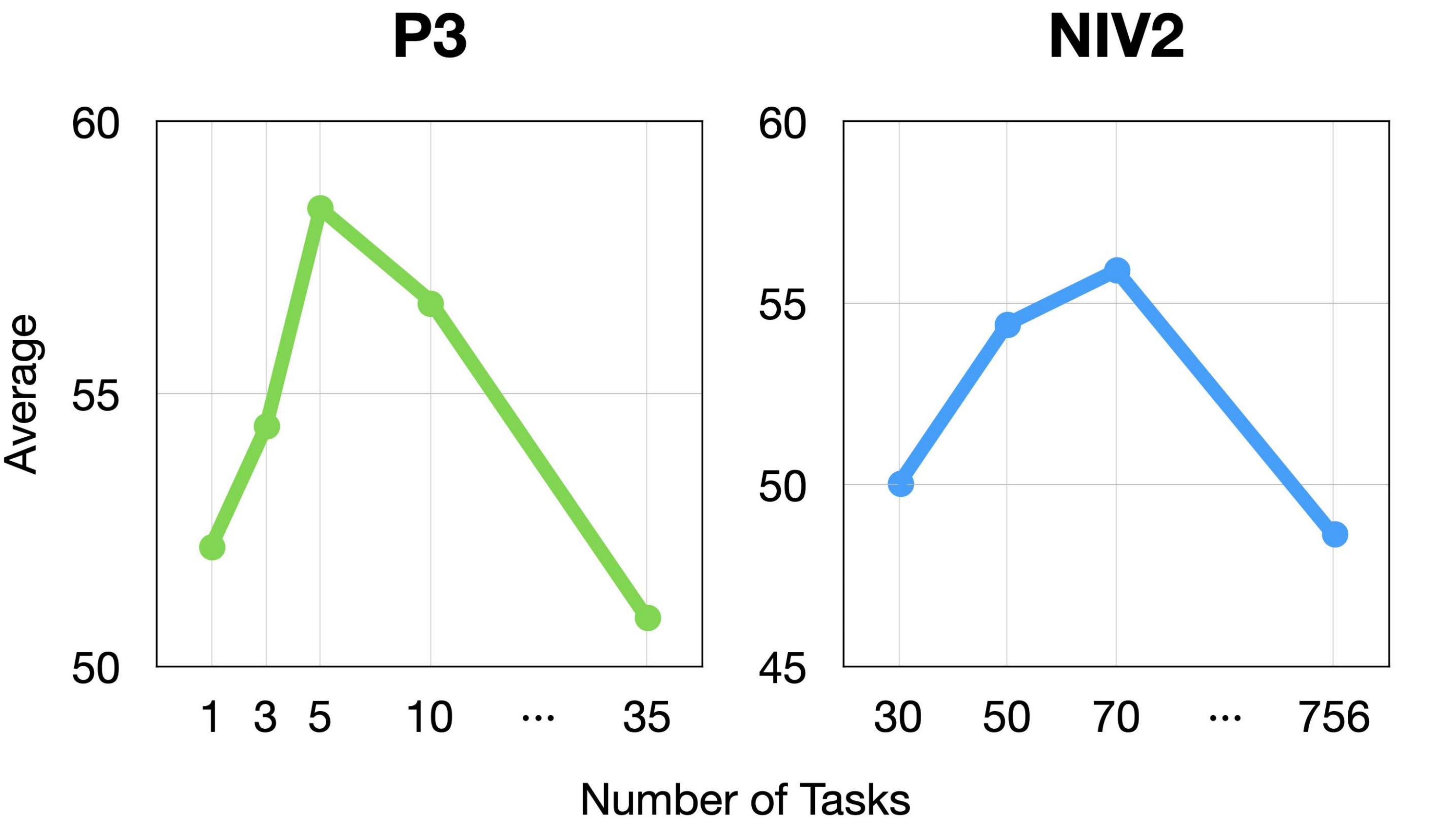}
\caption{Top-$k$ relevant task performance of P3 and NIV2 datasets. The left figure represents the average performance of 11 unseen tasks using T5(3B) + \textsc{InsTa}$_{\textbf{\textit{Aligned}-P3}}$, and the right figure represents the average performance of 33 unseen tasks using T5(3B) + \textsc{InsTa}$_{\textbf{\textit{Aligned}-NIV2}}$.}
\label{fig:topk_average}
\vspace{-3mm}
\end{figure}
\subsection{Scaling Relevant Tasks}
\label{sec:temp-topk}
We explore how the model performance is affected by the number of selected tasks during instruction tuning. Figure \ref{fig:topk_average} illustrates the average performance of T5(3B) + \textsc{InsTa}$_{\textbf{\textit{Aligned}-P3}}$ and T5(3B) + \textsc{InsTa}$_{\textbf{\textit{Aligned}-NIV2}}$, as the number of selected tasks ($k$) increases. The performance of the top-35 tasks for P3 and top-756 tasks for NIV2 corresponds to the fully instruction-tuned models, representing the scores of T0-3B and \textsc{T\textit{k}-Instruct-3B}, respectively. On average, we observe a progressive improvement in performance up to five tasks for P3 and seventy tasks for NIV2, after which it declines, affirming the adverse impact of non-relevant tasks during training. For more analysis and detailed results, please refer to Appendix \ref{sec:topk_analysis}.

\section{Conclusion}
In this study, we discover that selecting informative tasks for instruction tuning can be effectively achieved by exclusively using task instructions. Our experiments reveal that this method, particularly when aligned with the meta-dataset's instructions, surpasses traditional methods which depend on data samples to determine task relevance. Beyond its robust performance, our approach's most significant advantage is its simplicity: as long as the instruction (task description) of the target task can be described, our method can be applied. This approach marks a significant shift away from traditional methods that require exhaustive pairwise comparisons or the labor-intensive creation of data samples for new tasks. By adopting an \textit{instruction-only} strategy, our method simplifies and automates the task selection process for instruction tuning, providing a more efficient and practical approach to developing models in real-world scenarios.

\section*{Limitations}
While highlighting the effective task selection method in instruction tuning, we do not perform experimental results over the different sizes of model parameters other than the T5(3B) parameter model due to computational cost. For example, a language model bigger than 11B parameters may be less susceptible to negative transfer due to model capacity, or the effectiveness of our task selection method might stand out even more. Moreover, we only use the encoder-decoder architecture model in the paper. We leave the investigation on the decoder model, such as LLaMA 7B/13B, to our future work.

In this study, our focus is on two prominent meta-datasets, P3 and NIV2. Nonetheless, it is worth noting the existence of various other instruction tuning meta-datasets, such as FLAN-T5 \cite{Flan_t5} and the CoT (reasoning) collections \cite{cot_collection}. Extending our methodology to incorporate these additional meta-datasets will be an intriguing prospect for future research and exploration.

\section*{Acknowledgements}
We would like to express sincere thanks to Dongkyu Lee and Joel Jang for their thoughtful feedback on the paper.

\bibliography{anthology,custom}

\appendix
\label{sec:appendix}

\section{Previous Task Selection Methods in NLP}
\label{sec:previous_method}
Task selection is actively researched in two main fields. One is task selection from the perspective of intermediate-task transfer learning, and the other is task selection within the realm of instruction tuning, which is the focus of our research.

\subsection{Task selection in Intermediate-Task Transfer Learning}
Within the scope of intermediate-task transfer learning, the goal of task selection is to enhance the performance by further training the model on related intermediate tasks before fine-tuning it on a target task \cite{stilt,vu_transfer1,kung_transfer,poth_transfer}. Therefore, it presupposes the availability of a labeled dataset for the target task.

\citet{vu_transfer1} use BERT as a feature extractor to identify related tasks by comparing embeddings with an auxiliary task pool. In contrast, \citet{kung_transfer} develop a more efficient method, training a task discriminator with just 500 samples from each dataset, thereby reducing data needs. However, creating 500 labeled samples for a target task can still be burdensome, especially in an instruction-tuning setting. Moreover, in line with \citet{vu_transfer1}, this approach entails a high computational cost for training task discriminator and inference on all auxiliary instances.

\citet{poth_transfer} opt for a different approach, directly measuring pairwise transfer for all datasets instead of using samples' embedding. To reduce computational costs, they employ adapter structures. Nonetheless, this method still incurs significant costs in terms of target data construction and computational overhead because it requires training and inference across all models.

\begin{table*}[h!]
    \resizebox{\textwidth}{!}{\begin{tabular}{lccccccc}
    \toprule
    & \textbf{Task} & & \textbf{Target} & \textbf{Needed Target} & \textbf{Target Data} & & \textbf{Model} \\
    \textbf{Research} & \textbf{Selection} & \textbf{Perspective} & \textbf{Data} & \textbf{Data} & \textbf{Labels} & \textbf{Selector Training} & \textbf{Training \&} \\
    & \textbf{Method} & & \textbf{Needed} & \textbf{Samples} & \textbf{Required} & & \textbf{Inference} \\
    \midrule
    \multirow{2}{*}{\citet{vu_transfer1}} & \multirow{2}{*}{Sample-based} & Intermediate-task & \multirow{2}{*}{Yes} & \multirow{2}{*}{All} & \multirow{2}{*}{Yes} & All samples in & \multirow{2}{*}{No} \\
    & & transfer learning & & & & target task & \\
    \midrule
    \multirow{2}{*}{\citet{kung_transfer}} & \multirow{2}{*}{Sample-based} & Intermediate-task & \multirow{2}{*}{Yes} & \multirow{2}{*}{500 samples} & \multirow{2}{*}{Yes} & 500 samples for & \multirow{2}{*}{No} \\
    & & transfer learning & & & & all task (Total 18000) & \\
    \midrule
    \multirow{2}{*}{\citet{poth_transfer}} & Measaure all & Intermediate-task & \multirow{2}{*}{Yes} & \multirow{2}{*}{All} & \multirow{2}{*}{Yes} & \multirow{2}{*}{No} & All task \\
    & task pair & transfer learning & & & & & needed \\
    \midrule
    \multirow{2}{*}{\citet{ivison_transfer}} & \multirow{2}{*}{Sample-based} & \multirow{2}{*}{Instruction tuning} & \multirow{2}{*}{No} & 1000 samples & \multirow{2}{*}{No} & \multirow{2}{*}{No} & \multirow{2}{*}{No} \\
    & & & & (1 instruction) & & & \\
    \midrule
    \multirow{2}{*}{\textbf{\citet{roe}}} & \multirow{2}{*}{Sample-based} & \multirow{2}{*}{Instruction tuning} & \multirow{2}{*}{No} & 32 samples & \multirow{2}{*}{No} & \multirow{2}{*}{No} & \multirow{2}{*}{No} \\
    & & & & (10 instructions) & & & \\
    \midrule
    \multirow{2}{*}{\textbf{\citet{not_all_task}}} & Measaure all & \multirow{2}{*}{Instruction tuning} & \multirow{2}{*}{No} & \multirow{2}{*}{All} & \multirow{2}{*}{No} & \multirow{2}{*}{No} & All task \\
    & task pair & & & & & & needed \\
    \midrule
    \multirow{2}{*}{\citet{taskweb}} & Measaure all & \multirow{2}{*}{Instruction tuning} & \multirow{2}{*}{No} & \multirow{2}{*}{All} & \multirow{2}{*}{No} & \multirow{2}{*}{No} & All task \\
    & task pair & & & & & & needed \\
    \midrule
    \textsc{InsTa} & Instruction-based & Instruction tuning & No & (10 instructions) & No & No & No \\		
    \midrule
    & & & & & & 10 instructions for & \\
    \textsc{InsTa}$_{\textbf{\textit{Aligned}}}$ & Instruction-based & Instruction tuning & No & (10 instructions) & No &  training task $+$ 1500 STS & No \\
    & & & & & & samples (Total 1950) & \\
    \bottomrule
    \end{tabular}}
\caption{Characteristics and practicalities of task selection methodologies for the P3 meta-dataset in an instruction tuning setting. \textbf{Bold} indicates the studies used as baselines.}
\label{table:previous_methods_comparison}
\vspace{-3mm}
\end{table*}

\subsection{Task Selection in Instruction Tuning}
In task selection in instruction tuning, the objective is to select and train related tasks for improving the zero-shot performance of the target task. Unlike intermediate-task transfer learning, where samples consist solely of instances, data samples in instruction tuning include both instructions and instances. A model trained with such instructions can tackle new tasks when given new instructions, leveraging previously learned instructions as a basis.

\citet{ivison_transfer} propose a sample-based task selection methodology in instruction tuning. They encode all instances from a vast multitask data pool using a T5-3B model, then build a search index from the resulting representations. A key advantage of their approach is that there's no need to train an embedding generator with the target task's data samples, unlike task selection in intermediate-task transfer learning. However, the method still requires 1,000 unlabeled data samples from the target task. \citet{roe} adopt a method similar to \citet{ivison_transfer} to retrieve experts related to the target task for instruction tuning by encoding 32 data samples for each expert to calculate similarity. 

\citet{not_all_task,taskweb} apply a methodology similar to \citet{poth_transfer} in the context of instruction tuning, performing pairwise transfer across a variety of tasks. Notably, \citet{not_all_task} conduct pairwise transfer for all tasks in the P3 dataset, and these results are utilized as a baseline in their paper.

\subsection{Baseline for Task Selection in Instruction Tuning}
As noted by \citet{ivison_transfer}, task selection in intermediate-task transfer learning has several disadvantages compared to the task selection methodologies used in instruction tuning settings. These methodologies are primarily applied to classification tasks, require a large number of labeled samples for the target task, and involve high computational costs for training a model to generate task embeddings. For these reasons, we exclude these studies in our baselines. Additionally, we used \citet{roe}'s methodology as the baseline for the sample-based task selection instead of \citet{ivison_transfer} because \citet{roe}'s approach requires fewer data samples, making implementation easier. The characteristics and practical applicability of each methodology are summarized in the Table \ref{table:previous_methods_comparison}.

\begin{table}[t!]
    \resizebox{\columnwidth}{!}{\begin{tabular}{lcc}
    \toprule
    \textbf{Model} & \textbf{Time Complexity} & \textbf{P3 Avg.} \\
    \midrule
    \multirow{2}{*}{T5(3B) + PE \textsc{w/RoE}} & \( O((T_t + T_e) \cdot k \cdot n) \) & \multirow{2}{*}{53.26} \\
    & + \( O( T_t \cdot T_e \cdot k^2 \cdot n^2) \) & \\
    \multirow{2}{*}{T5(3B) + \textsc{InsTa}} & \( O((T_t + T_e) \cdot k) \) & \multirow{2}{*}{\textbf{55.71}} \\
    & + \( O(T_t \cdot T_e \cdot k^2) \) & \\
    \bottomrule
    \end{tabular}}
\caption{Comparison between T5(3B) + PE \textsc{w/RoE} and T5(3B) + \textsc{InsTa}. The top operand in the time complexity column represents the encoding complexity, while the bottom operand represents the similarity measurement complexity.}
\label{table:roe_vs_insta}
\vspace{-3mm}
\end{table}

\begin{table}[t!]
    \resizebox{\columnwidth}{!}{\begin{tabular}{lcc}
    \toprule
    \textbf{Model} & \textbf{Selection Time (Training)} & \textbf{P3 Avg.} \\
    \midrule
    T5(3B) + Pairwise Transfer & 35 * 32h & 57.86 \\
    T5(3B) + \textsc{InsTa}$_{\textbf{\textit{Aligned}}}$ & 5m & \textbf{57.97} \\
    \bottomrule
    \end{tabular}}
\caption{Comparison between T5(3B) + Pairwise Transfer and T5(3B) + \textsc{InsTa}$_{\textbf{\textit{Aligned}}}$. Note that the pairwise transfer approach takes considerable selection time since it individually trains on every training dataset.}
\label{table:pairwise_vs_insta_aligned}
\vspace{-3mm}
\end{table}

\section{Efficiency Analysis}
\label{sec:efficiency}
To evaluate the efficiency of our task selection method relative to other alternatives, we quantify it based on time and complexity. Specifically, we conduct a comparative analysis of the sample-based approach and the pairwise transfer approach using the same configurations and models as outlined in Table \ref{table:p3_selector}.

\paragraph{Sample-Based Selection vs. \textsc{InsTa}}  
Table \ref{table:roe_vs_insta} demonstrates the efficiency differences between the sample-based approach and our instruction-based approach. For this comparison, we utilize the sample-based T5(3B) + PE \textsc{w/RoE} and our instruction-based T5(3B) + \textsc{InsTa}. Both approaches employ SentenceBERT-based cosine similarity, enabling us to evaluate their efficiency by examining differences in time complexity. This time complexity incorporates both the sentence encoding process and the similarity computation. Let \( T_t \) represent the number of training tasks, \( T_e \) the number of evaluation tasks, \( k \) the average number of instructions, and \( n \) the number of data samples. The time complexity for encoding in the sample-based approach can be expressed as \( O((T_t + T_e) \cdot k \cdot n) \). In contrast, the instruction-based methodology, which does not require processing multiple data samples per instruction, has an encoding time complexity of \( O((T_t + T_e) \cdot k) \). The time complexity for similarity computation is also derived similarly. Considering all combinations of training and evaluation tasks, the complexity for the sample-based scenario is \( O(T_t \cdot T_e \cdot k^2 \cdot n^2) \), whereas for the instruction-based scenario, it is \( O(T_t \cdot T_e \cdot k^2) \). This indicates that our approach is at least 32 times faster than the sample-based T5(3B) + PE \textsc{w/RoE}, which requires 32 samples $(n=32)$.

\paragraph{Pairwise Transfer vs. \textsc{InsTa}$_{\textbf{\textit{Aligned}}}$}  
Table \ref{table:pairwise_vs_insta_aligned} presents the training time required for our method compared to the pairwise transfer method, which trains and evaluates every task pair. In this comparison, we analyze T5(3B) + Pairwise Transfer and T5(3B) + \textsc{InsTa}$_{\textbf{\textit{Aligned}}}$. Following the experimental setting on \citet{not_all_task}, with a batch size of 512 and 1000 steps, it takes approximately 32 hours on 1 A100 GPU for one task training. The P3 has 35 training datasets, and the total time required is approximately 35 * 32 hours. Conversely, our methodology takes about 5 minutes to train the SentenceBERT model on a single A100 GPU. This is only considering the training time, and the time difference becomes even greater when including inference time.

As indicated in Tables \ref{table:roe_vs_insta} and \ref{table:pairwise_vs_insta_aligned}, our task selection approach is overwhelmingly efficient and demonstrates robust performance. This facilitates the development of an optimized model for the specific target task in an instruction tuning setting.


\section{Addressing Potential Concerns}
\label{potential concerns}
\paragraph{Potential for Bias}
Our task selection process, which relies solely on instructions, might initially raise concerns about potential bias. If the training data selected is biased, the tasks chosen could perpetuate this bias in the fine-tuned LLM. However, it is crucial to note that our methodology robustly counters such biases for two key reasons. Firstly, many of the publicly available instruction tuning datasets are composed of diverse datasets, which are predominantly high-quality and unbiased. Secondly, even if some of the selected datasets exhibit bias, this issue is effectively mitigated by the majority of other selected unbiased datasets. For instance, in the NIV2 experiment, we used 70 out of 756 datasets for training. Even if some datasets contained biases, the impact is neutralized by training with the remaining datasets. Our experiments support this approach, as we confirmed no performance degradation due to bias across a total of 791 training datasets and 85 evaluation tasks (P3: 11, NIV2: 33, BB: 14, BBH: 27). These results demonstrate that our approach ensures robustness against potential biases.

\paragraph{Limited Generalizability}
Our task selection method relies on the quality and completeness of instructions for the target task. Therefore, there may be concerns that if the instructions are inaccurate, performance could deteriorate. However, it's important to note that our methodology can effectively operate with just one well-written instruction. Creating such an instruction is not typically burdensome. Moreover, should challenges arise in defining the instruction, they can be readily resolved using models like GPT-4 (please refer to Table \ref{table:niv2_query} for the GPT-4 query). For instance, for the BBH datasets, which initially lacked instructions, we utilized GPT-4 to generate one instruction per task for evaluation purposes. The results demonstrated significant performance improvements in most tasks (see Table \ref{table:bbh}). Overall, while our methodology cannot completely eliminate issues with instruction quality, these limitations do not severely restrict its generalizability and can be easily alleviated.

\section{Off-the-shelf Embedding Model}
\label{off-the-shelf embedding model}
We employ the Sentence Transformer \cite{sentence-bert} as an off-the-shelf embedding model in our study. Specifically, we utilize the "sentence-transformers/bert-large-nli-stsb-mean-tokens" checkpoint, which constitutes a 340 million parameter sentence-bert model trained on a combination of Natural Language Inference (NLI) data and the Semantic Textual Similarity (STS) benchmark data. To quantify the similarity between two given instructions, we measure the cosine similarity of their extracted embedding representations derived from this model.

We further train the Sentence Transformer using instructions from the meta-dataset to learn the style of the instructions. Besides the instructions, we sample 500 instances from MRPC, PAWS, and QQP datasets each and append them to training samples. These datasets are paraphrase identification datasets, ensuring that the model retains its generality throughout the training process.

During training, we employ a learning rate of 1e-6 for the P3 dataset and 1e-5 for the NIV2 dataset. The remaining training configurations align with those described in \citet{sentence-bert}. Specifically, we conduct training for five epochs and select the checkpoint corresponding to the best validation performance for subsequent analysis and evaluation.

\section{Training Details}
\label{training_setup}
Following \cite{T0}, we conduct our experiments in a held-out setting. Specifically, we exclude the task clusters of the evaluation tasks from the training dataset pool and then train our model, subsequently evaluating it on the excluded held-out tasks. For instance, in the case of the P3 dataset, we remove task clusters such as sentence completion, NLI, coreference resolution, and word sense disambiguation from the training set and train only on the selected tasks from the remaining datasets. It ensures that the evaluation set remains entirely unseen, thereby achieving a true zero-shot setting.

Additionally, SentenceBert training for \textsc{InsTa}$_{\textbf{\textit{Aligned}}}$ follows the same principle. We exclude all instructions from the evaluation clusters for task selector training and train using only instructions from the training dataset pool, thus preventing data contamination issues.

We truncate input and target sequences to 768 and 256 tokens, respectively. We train all models with a batch size of 256 using Adafactor Optimizer for both P3 and NIV2 instruction tuning. We train the model using 16 NVIDIA A100 GPUs (each with 40GB). Each training for top-$k$ P3 and NIV2 instruction tuning requires less than 1 hour per epoch, and ends up 2-3 hours for 3 epochs.

Note that P3 and NIV2 meta-datasets are license free for research purpose and open-sourced with the code.

\section{Examples and Formulation of Instructions of P3 and NIV2.}
\subsection{Examples of P3 and NIV2 Instructions.}
\label{sec:examples_of_p3_niv2_instruction}
\citet{T0} introduces natural language prompts for all datasets to enable zero-shot experimentation. They name it as prompt, template, and instruction, and we only use term instruction for this paper. The instruction they define consists of an input template and a target template, along with a collection of associated meta-data. The instructions are in functions mapping a data instance into natural language for the input and target sequences. For example, in the case of an NLI dataset, the example includes fields for \textsc{Premise, Hypothesis, Label}, and input instruction would be 

\begin{boxA}
If \{\{Premise\}\} is true, is it also true that \{\{Hypothesis\}\}? 
\end{boxA}

The target instruction can be defined with the label choices \textsc{\{\{Choices[label]\}\}} but we only used input instruction without replacing the placeholder with actual data instance to find similar tasks.

In contrast with P3, NIV2 defines Instruction schema and divides the component as \textsc{Definition, Positive Examples, Negative Examples}. \textsc{Definition} defins a given task in natural language of how input text is expected to be mapped to an output text. Each task has a single \textsc{Definition}, and we only use \textsc{Definition} part as an instruction for each task and utilize it for task selection. The example below is \textsc{Definition} part of task1640, an \textsc{Answerability Classification} task. Note that, unlike P3 instruction, NIV2 instruction comprises of natural language without any placeholder.

\begin{boxA}
Given a paragraph from a wikipedia article about some topic, and a question related to the topic, determine whether
the question is answerable from the paragraph. If the question is answerable, answer “True”, otherwise, answer
“False”.
\end{boxA}

\subsection{P3 Instruction Formulation}
\label{sec:p3_instruction_formulation}
The original P3 instructions contained unique placeholders, which could potentially act as misleading shortcuts. In order to preserve the integrity of the instruction's original meaning, we have replaced these placeholders with the terms "\{\{text\}\}" and "\{\{candidate\}\}". In the classification task, if the options are provided within the instruction, we substitute them with "candidate"; otherwise, we use "text" for all replacements. Since we utilize only the input instruction and not the output label, we recommend examining the examples in the Table \ref{table:p3_instruction_examples} to understand the appearance of the original P3 instruction and how it has been modified.

\subsection{NIV2 Input Formats}
\label{niv2_input_formats}
\begin{figure}[b!]
\centering
    \includegraphics[width=0.8\columnwidth]{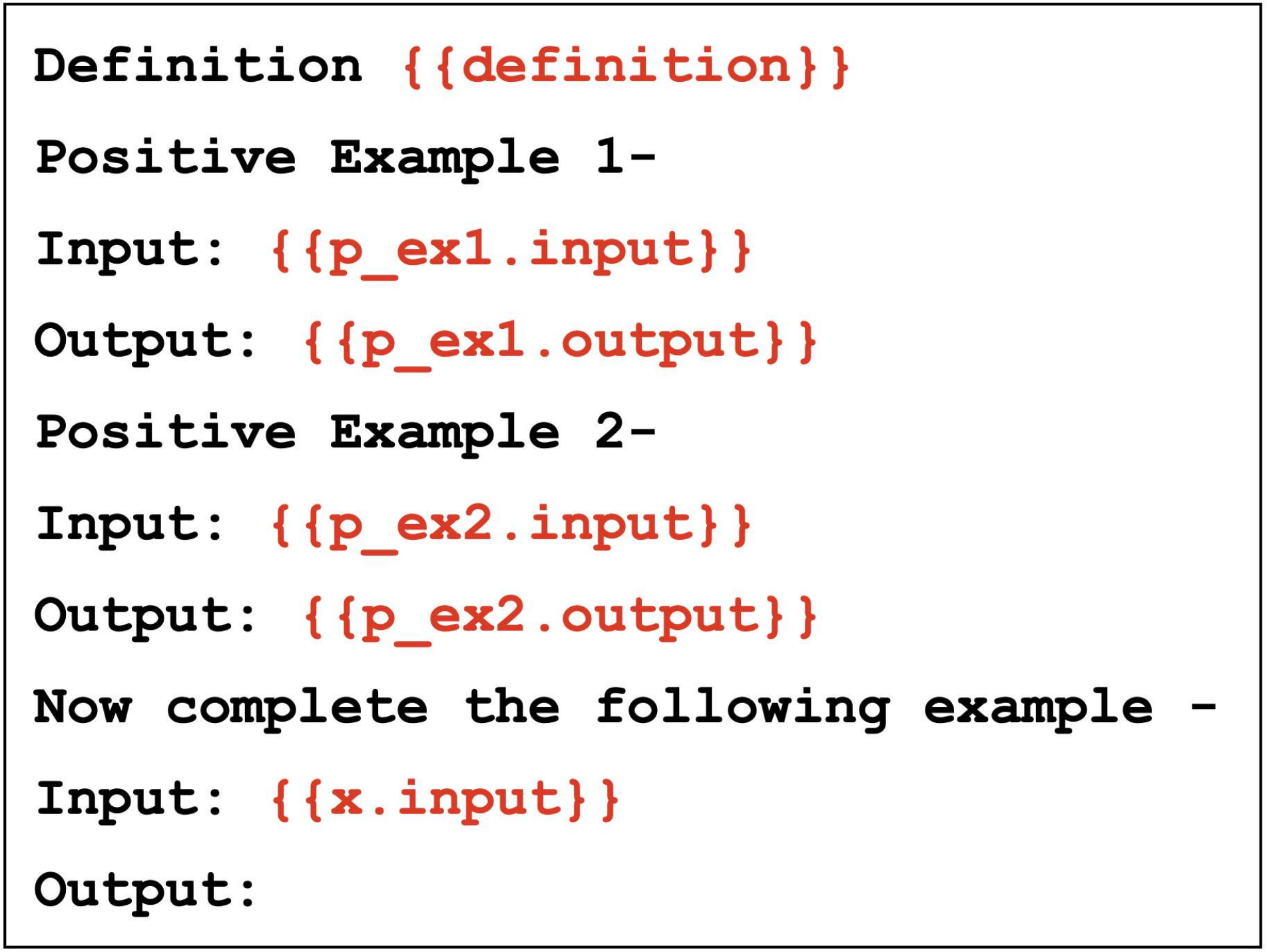}
\caption{Input encoding for [\textit{Def + Pos(2)}] setting.}
\label{fig:def_pos}
\vspace{-1mm}
\end{figure}

\begin{figure}[b!]
\centering
    \includegraphics[width=0.8\columnwidth]{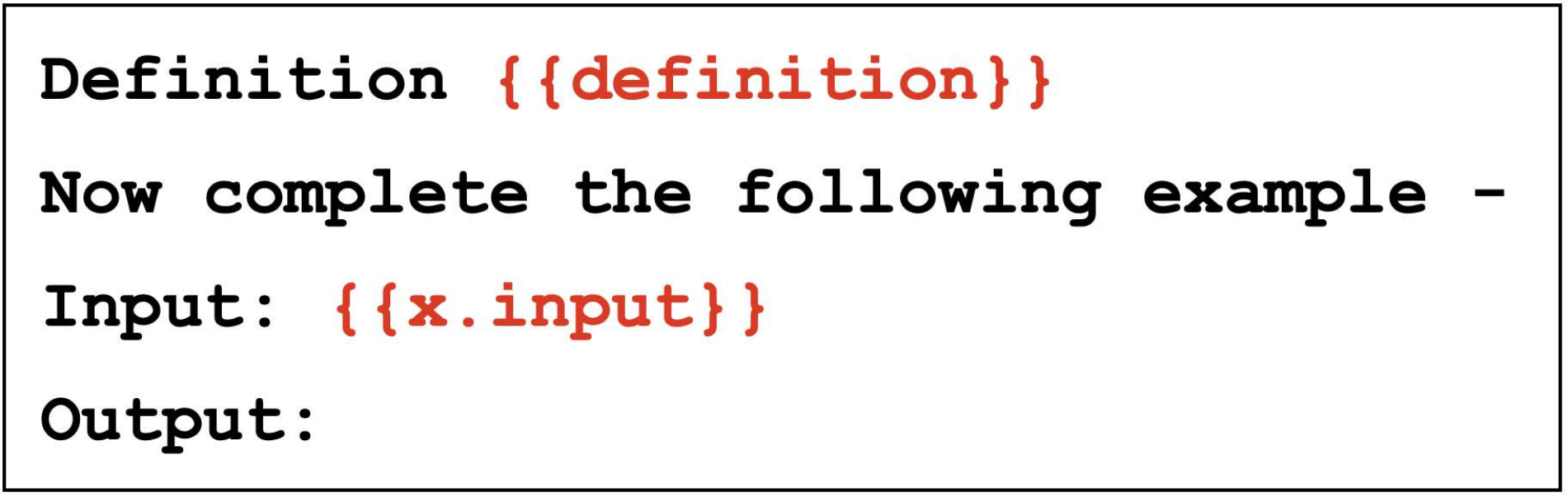}
\caption{Input encoding for [\textit{Def}] setting.}
\label{fig:def}
\vspace{-3mm}
\end{figure}

\citet{supernatural} demonstrate various instruction composition using \textit{Task Definition}, \textit{Positive Task Examples}, \textit{Negative Task Examples}, and \textit{Explanation}. Out of various composition settings, we use [\textit{Def + Pos(2)}] setting for Section \ref{sec:NIV2 Results} and [\textit{Def}] setting for Section \ref{sec:BBH Results}. Figure \ref{fig:def_pos} and Figure \ref{fig:def} represent input encoding for the above settings, respectively.

\begin{table}[h!]
    \resizebox{\columnwidth}{!}{\begin{tabular}{lc}
    \toprule
    \textbf{Task} & \textbf{Instruction} \\
    \midrule
    RTE & Suppose \{\{premise\}\} Can we infer that \\
    & "\{\{hypothesis\}\}"? Yes or no? \\
    & $\downarrow$ \\
    & Suppose \{\{text\}\} Can we infer that \\
    & "\{\{text\}\}"? Yes or no? \\
    \midrule
    Amazon & Title: \{\{title\}\}\textbackslash nProduct review: \{\{text\}\}\textbackslash n \\
    & Would you say this review depicts the product in a \\
    & \{\{choices[1]\}\} or \{\{choices[0]\}\} light?\textbackslash n \\
    & $\downarrow$ \\
    & Title: \{\{text\}\}\textbackslash nProduct review: \{\{text\}\}\textbackslash n \\
    & Would you say this review depicts the product in a \\
    & \{\{candidate\}\} or \{\{candidate\}\} light?\textbackslash n \\
    \midrule
    Winogrande & \{\{text\}\}\textbackslash nReplace the \_ in the above \\
    & sentence with the correct option: \textbackslash n\\
    & - \{\{choices[0]\}\}\textbackslash n- \{\{choices[1]\}\}\textbackslash n \\
    & $\downarrow$ \\
    & \{\{text\}\}\textbackslash nReplace the \_ in the above \\
    & sentence with the correct option: \textbackslash n\\
    & - \{\{candidate\}\}\textbackslash n- \{\{candidate\}\}\textbackslash n \\
    \midrule
    QuaRel & Here's a short story: \{\{question\}\}.\textbackslash n\textbackslash nWhat is \\
    & the most sensical answer between "\{\{choices[0]\}\}" \\
    & and  "\{\{choices[1]\}\}"?\textbackslash n \\
    & $\downarrow$ \\
    & Here's a short story: \{\{text\}\}.\textbackslash n\textbackslash nWhat is \\
    & the most sensical answer between "\{\{candidate\}\}" \\
    & and  "\{\{candidate\}\}"?\textbackslash n \\
    \midrule
    Wiki Bio & Facts:\textbackslash n\{\{concepts\}\}\textbackslash nBased on these \\
    & bullet points, write a short biography \\
    & describing the life of \{\{person\}\}. \\
    & $\downarrow$ \\
    & Facts:\textbackslash n\{\{text\}\}\textbackslash nBased on these \\
    & bullet points, write a short biography \\
    & describing the life of \{\{text\}\}. \\
    \midrule
    PAWS & Sentence 1: \{\{sentence1\}\}\textbackslash nSentence 2: \{\{sentence2\}\} \\
    & \textbackslash nQuestion: Do Sentence 1 and Sentence 2 express \\
    & the same meaning? Yes or No? \textbackslash n \\
    & $\downarrow$ \\
    & Sentence 1: \{\{text\}\}\textbackslash nSentence 2: \{\{text\}\} \\
    & \textbackslash nQuestion: Do Sentence 1 and Sentence 2 express \\
    & the same meaning? Yes or No? \textbackslash n \\
    \midrule
    MultiNews & Write a summary of the following articles:\textbackslash n\textbackslash n \\
    & Document: \{\{text\}\}\textbackslash n \\
    & $\downarrow$ \\
    & Write a summary of the following articles:\textbackslash n\textbackslash n \\
    & Document: \{\{text\}\}\textbackslash n \\
    \bottomrule
    \end{tabular}}
\caption{Examples of P3 instruction formulation. The unique values in placeholder are unified using \{\{text\}\} and \{\{candidate\}\}. Note that we use unified instructions to train the selector model.}
\label{table:p3_instruction_examples}
\vspace{-3mm}
\end{table}

\section{GPT-4 API Generation}
\subsection{NIV2 Instruction Generation}
\label{sec:niv2_instruction_generation}
To reproduce positive example for NIV2 instructions, we use GPT-4 API to generate similar instruction to original ones. Table \ref{table:niv2_query} represents API query used for instruction generation, and Table \ref{table:niv2_instruction} represents examples of generated instruction from GPT-4.

\subsection{Big-Bench Hard Instruction Generation}
\label{sec:bbh_generation}
The BBH tasks don't have refined \textit{Task Definition} like NIV2 datasets. In order to select relevant tasks using NIV2 datasets, and to compare the performance with \textsc{T\textit{k}-Instruct}, we generate instruction for BBH tasks using GPT-4. Table \ref{table:bbh_query} represents API query used for instruction generation, and Table \ref{table:bbh_instructions_1}, \ref{table:bbh_instructions_2} and \ref{table:bbh_instructions_3} show generated 27 instructions for each task.

\section{Pairwise Transfer Task Selection vs. Instruction-Based Task Selection}
\label{pairwise and i-bts selected choice}
\begin{figure*}[ht!]
\centering
    \includegraphics[width=\textwidth]{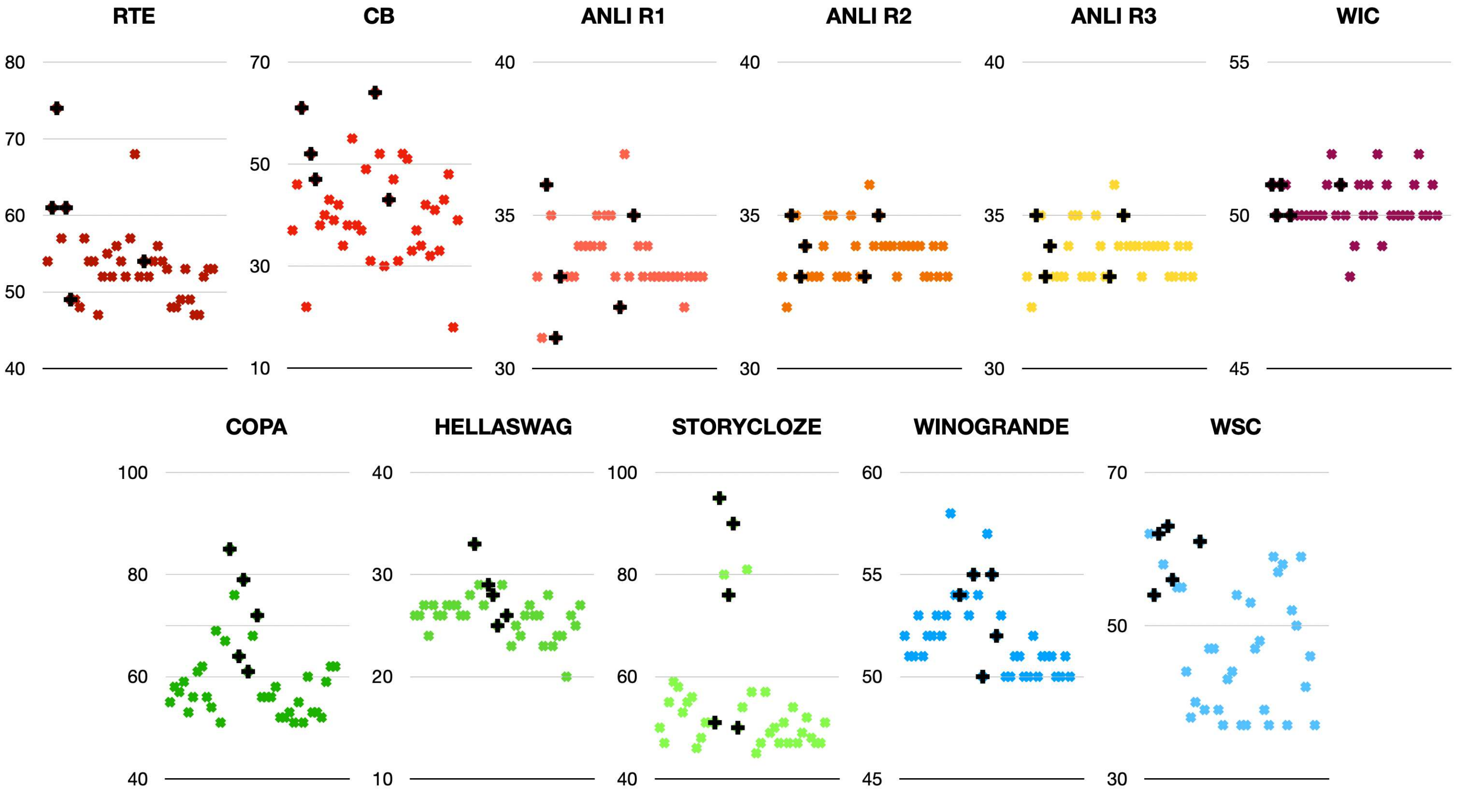}
\caption{Pairwise transferability result datasets and top-5 selected tasks from T5(3B) + \textsc{InsTa}$_{\textbf{\textit{Aligned}-P3}}$. The y-axis represents similarity scores for pairwise transferability scores, and tasks selected by T5(3B) + \textsc{InsTa}$_{\textbf{\textit{Aligned}-P3}}$ are marked as black cross.}
\label{fig:pairwise_diagram}
\vspace{-3mm}
\end{figure*}

\citet{not_all_task} measure pairwise transfer relationships on T5(3B) in all task pair of P3 meta-dataset. They measure the value of its average score on different instructions for every dataset, and they include instructions only related to the original tasks for evaluation. The T5(3B) + Pairwise Transfer model in Figure \ref{fig:pairwise_diagram} selects top-5 tasks that scored the highest transferability per target tasks. Figure \ref{fig:pairwise_diagram} represents top-5 selected tasks from T5(3B) + \textsc{InsTa}$_{\textbf{\textit{Aligned}-P3}}$ and all the transferability scores. It can be verified that T5(3B) + \textsc{InsTa}$_{\textbf{\textit{Aligned}-P3}}$ selects relevant tasks in accordance with pairwise transferability result in many tasks.

\section{Experimental Detail for Data Sample-based Task Selection}
\label{sec:data_sample_based_experimental_detail}
In our experiments, we adopt the data sample-based task selection approach as detailed in the experimental settings of \citet{selector_ye,roe}. To ascertain the similarity between tasks, previous works randomly sample 32 instances from each dataset, which are then paired with the corresponding instructions. Following this, embedding information are extracted using a dense retriever, and cosine similarities are calculated through matrix multiplication between the embedding vectors of the training and target tasks. Consistent with prior studies, we select 32 instances for all instructions in our training datasets, measure the similarity, and prioritize tasks that feature the highest scoring instructions. Apart from the task selection methodology, the instruction tuning procedures remain identical to those used in training the T5(3B) + \textsc{InsTa} model.

\section{Analysis on Selecting Top-\textit{k} Relevant Tasks}
\label{sec:topk_analysis}
\begin{figure*}[ht!]
\centering
    \includegraphics[width=\textwidth]{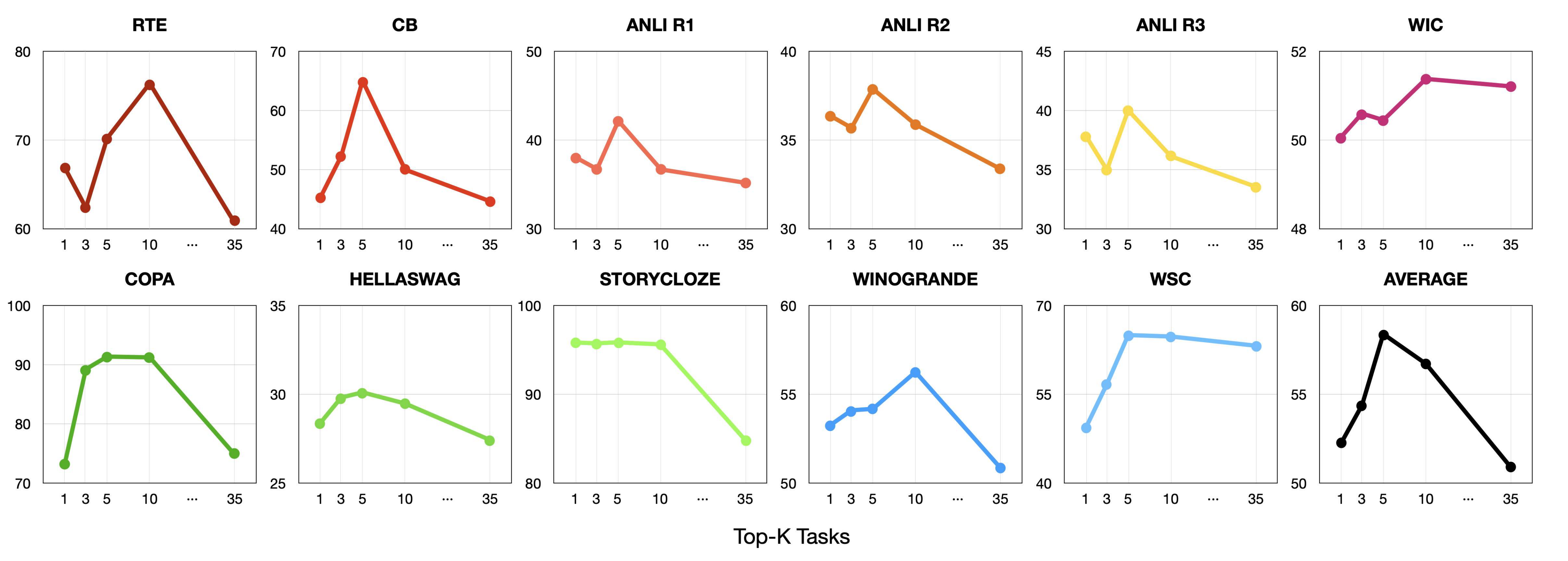}
\caption{Top-$k$ relevant task performance of P3 datasets. The connected line represents the performance of T5(3B) + \textsc{InsTa}$_{\textbf{\textit{Aligned}-P3}}$, and note that the score of top-35 tasks represents the performance of T0-3B. }
\label{fig:topk}
\vspace{-3mm}
\end{figure*}

In our experiments, we utilize the top-\textit{k} tasks selection approach, following previous work \cite{taskweb, not_all_task}, while acknowledging its potential risks. This method may occasionally include tasks with marginal relevance or exclude significant ones. Figure \ref{fig:topk} presents the performance of 11 unseen datasets in P3 as \textit{k} varies. The experimental results show that training the model with all 35 tasks reveals a performance, but note that the optimal \textit{k} for each dataset varies across datasets. WiC and Winogrande demonstrate optimal results when the \textit{k} is 10, while other datasets perform best when the \textit{k} is 5. 

Although we have attempted to use the selector's probability score to determine the cutoff, variations in the data used for training the selector and the type of models that could serve as selectors have introduced inconsistencies. These issues are recognized but beyond the scope of this study. Thus, we do not address this directly in our research and intend to investigate more precise measures of task relevance in future work.

\section{Scaling The Number of Instructions in Alignment}
\label{sec:scale_instruction}
Table \ref{table:scale_instruction} shows how performance changes in the P3 meta-dataset based on the number of instructions learned during task selection. The same instructions are used across task selection, model training, and inference, but what varies is the number of instructions used for \textsc{InsTa} alignment. For training selector model using instructions for alignment, we scale the number of instructions by average of 2, 4, and 6.54, and 6.54 represents the aligned model \textsc{InsTa}$_{\textbf{\textit{Aligned}-P3}}$ used in the entire paper.

Experimental results demonstrate an increase in the robustness of task selection as the number of instructions increases. We hypothesize that by learning a variety of instructions, the task selector becomes proficient in handling different styles of instruction formats. This enables it to effectively identify relevant tasks when presented with various formats of instructions for new tasks.

\section{Learning Different Formats of Meta Datasets and Its Impact on Performance}
\label{sec:diff_meta_dataset}
In the previous section, we explore how task selection performance varies with the number of instructions in the same meta dataset. In this section, we conduct experiments to see if learning meta datasets with different instruction formats together would lead to performance degradation compared to learning each meta dataset separately. Performance is measured on the held-out tasks of the P3 dataset, and the T5(3B) + \textsc{InsTa}$_{\textbf{\textit{Aligned}-(P3+NIV2)}}$ is further aligned using \textsc{InsTa} on instructions from both the P3 and NIV2 datasets. 

Our results, summarized in Table \ref{table:insta_p3_niv2_combined}, indicate that the average performance T5(3B) + \textsc{InsTa}$_{\textbf{\textit{Aligned}-(P3+NIV2)}}$ of is on par with that of P3 alone, with some tasks even showing improved results. This evidence suggests that integrating the unique instructional style of the NIV2 dataset does not substantially impair the task selection capabilities in P3. It implies that training the task selector on diverse meta-datasets enables it to adapt to a broader range of instruction formats and tasks without declining performance.

\begin{table*}[h!]
    \resizebox{\textwidth}{!}{\begin{tabular}{lcccccccccccc}
    \toprule
    \multirow{2}{*}{\textbf{Method}} & \multicolumn{5}{c}{NLI} & \multicolumn{3}{c}{Sentence Completion} & \multicolumn{2}{c}{Coref. Resol.} & WSD & \multirow{2}{*}{\textbf{Total Avg.}}
    \\ \cmidrule(lr){2-6} \cmidrule(lr){7-9} \cmidrule(lr){10-11} \cmidrule(lr){12-12} & \textbf{RTE} & \textbf{CB} & \textbf{AN. R1} & \textbf{AN. R2} & \textbf{AN. R3} & \textbf{COPA} & \textbf{Hellasw.} & \textbf{StoryC.} & \textbf{Winogr.} & \textbf{WSC} & \textbf{WiC} &  \\
    \midrule
    T5(3B) + \textsc{InsTa} & 73.86 & 55.10 & 36.82 & 34.77 & 35.27 & 91.00 & 27.63 & \underline{94.10} & 55.26 & 56.13 & \textbf{52.84} & 55.70 \\  
    \midrule
    T5(3B) + \textsc{InsTa}$_{\textbf{\textit{Aligned}-P3(i=2)}}$ & \underline{77.51} & \underline{57.91} & 37.80 & 36.19 & 37.07 & \underline{91.50} & \underline{30.62} & 91.69 & 55.12 & \underline{56.85} & \underline{52.02} & 56.75 \\
    T5(3B) + \textsc{InsTa}$_{\textbf{\textit{Aligned}-P3(i=4)}}$ & 75.67 & \textbf{63.27} & \textbf{40.21} & \textbf{37.47} & \textbf{39.17} & 89.25 & 27.74 & 92.78 & \underline{55.58} & 53 & 50.49 & \underline{56.78} \\
    T5(3B) + \textsc{InsTa}$_{\textbf{\textit{Aligned}-P3}}$ & \textbf{77.87} & 56.89 & \underline{38.28} & \underline{36.30} & \underline{37.18} & \textbf{92.50} & \textbf{31.40} & \textbf{95.86} & \textbf{56.37} & \textbf{64.42} & 50.61 & \textbf{57.97} \\
    \bottomrule
    \end{tabular}}
\caption{Performance of the T5(3B) + \textsc{InsTa}$_{\textbf{\textit{Aligned}-P3}}$ as the number of instructions used for selector training varies. The expression \textbf{(i=N)} indicates a number of instructions used for the alignment process. Note that our T5(3B) + \textsc{InsTa}$_{\textbf{\textit{Aligned}-P3}}$ uses 6.54 instructions on average. The best comparable performances are \textbf{bolded} and second best \underline{underlined}.}
\label{table:scale_instruction}
\vspace{-3mm}
\end{table*}

\begin{table*}[h!]
    \resizebox{\textwidth}{!}{\begin{tabular}{lcccccccccccc}
    \toprule
    \multirow{2}{*}{\textbf{Method}} & \multicolumn{5}{c}{NLI} & \multicolumn{3}{c}{Sentence Completion} & \multicolumn{2}{c}{Coref. Resol.} & WSD & \multirow{2}{*}{\textbf{Total Avg.}}
    \\ \cmidrule(lr){2-6} \cmidrule(lr){7-9} \cmidrule(lr){10-11} \cmidrule(lr){12-12} & \textbf{RTE} & \textbf{CB} & \textbf{AN. R1} & \textbf{AN. R2} & \textbf{AN. R3} & \textbf{COPA} & \textbf{Hellasw.} & \textbf{StoryC.} & \textbf{Winogr.} & \textbf{WSC} & \textbf{WiC} &  \\
    \midrule
    T0-3B & 60.61 & 44.64 & 35.17 & 33.37 & 33.55 & 74.75 & 27.42 & 84.82 & 50.84 & \underline{63.22} & \underline{51.21} & 50.87 \\
    T5(3B) + \textsc{InsTa}$_{\textbf{\textit{Aligned}-P3}}$ & \textbf{77.87} & \underline{56.89} & \underline{38.28} & \underline{36.30} & \underline{37.18} & \textbf{92.50} & \textbf{31.40} & \textbf{95.86} & \textbf{56.37} & \textbf{64.42} & 50.61 & \textbf{57.97} \\
    T5(3B) + \textsc{InsTa}$_{\textbf{\textit{Aligned}-(P3+NIV2)}}$ & \underline{72.17} & \textbf{57.65} & \textbf{40.67} & \textbf{38.35} & \textbf{38.63} & \underline{91.50} & \underline{29.32} & \underline{95.00} & \underline{55.62} & 63.10 & \textbf{51.27} & \underline{57.57} \\
    \bottomrule
    \end{tabular}}
\caption{Comparison between performance of \textsc{InsTa}$_{\textbf{\textit{Aligned}}}$ when the selector trained with P3 instructions only (T5(3B) + \textsc{InsTa}$_{\textbf{\textit{Aligned}-P3}}$), and with combination of P3 and NIV2 instructions (T5(3B) + \textsc{InsTa}$_{\textbf{\textit{Aligned}-(P3+NIV2)}}$). The best comparable performances are \textbf{bolded} and second best \underline{underlined}.}
\label{table:insta_p3_niv2_combined}
\vspace{-3mm}
\end{table*}

\begin{table*}[h!]
    \resizebox{\textwidth}{!}{\begin{tabular}{lc}
    \toprule
    \textbf{GPT-4 Query} & \\
    \midrule
    You have to paraphrase definition of task when one definition of the task is given. Make sure you do not mention type of given task in definition. & \\
    Make a similar definitions without repetition. Separate each definition by two newline character.& \\
    & \\
    Example - & \\
    Task : Answerability Classification & \\
    Definition 1 : Given a paragraph from a Wikipedia article about some topic, and a question related to the topic, determine whether the question  & \\
    is answerable from the paragraph. If the question is answerable, answer "True", otherwise, answer "False". & \\
    Definition 2 : In this task you will be given a question and a passage. You need to determine if the answer to the question is contained in the passage.  & \\
    If the answer can be found in the passage you should output 'True'. If the answer cannot be found in the passage you should output 'False'. & \\
    & \\
    Read the examples above and generate similar task definition for given task type and definition. & \\
    \bottomrule
    \end{tabular}}
\caption{GPT-4 query used for NIV2 instruction generation.}
\label{table:niv2_query}
\end{table*}

\begin{table*}[h!]
    \resizebox{\textwidth}{!}{\begin{tabular}{lc}
    \toprule
    \multicolumn{2}{c}{\textbf{Answer Verification \textsc{Task846}}} \\
    \midrule
    Original & Given a passage with a question and an answer to that question, classify if the answer actually answers the question\\
    & into 'yes' or 'no'. Output 'yes' if the answer answers the question and output 'no' if the answer does not\\
    & answer the question.\\ 
    \midrule
    GPT-Generated & For this activity, you will be presented with a passage, alongside a question and a proposed answer.\\
    & Your objective is to evaluate whether the provided answer satisfactorily addresses the question. If it does,\\
    & you should indicate this with a 'yes'; if it does not, your response should be 'no'. \\
    \midrule
    \multicolumn{2}{c}{\textbf{Program Execution \textsc{Task850}}} \\
    \midrule
    Original & In this task you will be given a string and you should find the longest substring that is a palindrome. \\
    & A palindrome is a string that is the same backwards as it is forwards. If the shortest possible palindrome\\
    & is length 1 you should return the first character. \\
    \midrule
    GPT-Generated & In the given activity, the objective is to analyze a sequence of characters and identify the most \\
    & extensive segment within it that reads identically from both ends. Should the minimal length for such \\
    & a segment be a single character, the initial character of the sequence should be returned as the result. \\
    \midrule
    \multicolumn{2}{c}{\textbf{Text Categorization \textsc{Task681}}} \\
    \midrule
    Original & Given a comment text in Malayalam, classify the comment into one of these categories (i) Hope speech, \\
    & (ii) Not Hope Speech or (iii) Not in Expected Language. A hope speech contains content that is \\
    & encouraging, positive or supportive contents and talks about equality, diversity or inclusion \\
    \midrule
    GPT-Generated & In this activity, you will be presented with a piece of text in Malayalam. Your objective is to assign the text to \\
    & one of the following classifications: (i) Hopeful discourse, which includes messages that are uplifting, \\
    & affirmative, or bolstering, and often pertains to themes of unity, variety, or acceptance; \\
    & (ii) Non-hopeful discourse, or (iii) Text that is not composed in the anticipated language. \\
    \bottomrule
    \end{tabular}}
\caption{Examples of NIV2 task instruction generated by GPT-4.}
\label{table:niv2_instruction}
\end{table*}

\begin{table*}[h!]
    \resizebox{\textwidth}{!}{\begin{tabular}{lc}
    \toprule
    \textbf{GPT-4 Query} & \\
    \midrule
    You have to generate definition of the given task. You will be given two examples and each example will have a instance and definition. & \\
    & \\
    Example 1 - & \\
    Task : sentiment analysis & \\
    Input: Tweet: @KimWalshUk aw poor sarah  shouldn't it be Cheryl upset cause it's in Newcastle isn't it lol? Question: is it a negative tweet? & \\
    Output : yes & \\
    Definition : In this task, you are given a text from tweets and a boolean question whether this tweet has positive sentiment or negative & \\
    sentiment. Your task is to generate answer ""yes"" when the tweet  & \\
    has that particular sentiment, otherwise generate answer ""no"". & \\
     & \\
    Example 2 -  & \\
    Task : question answering & \\
    Input: What is a place that is far away from your house and where you could consume beer? & \\
    (A)refrigerator (B)friend's house (C)keg (D)neighbor's house (E)kitchen & \\
    Output : B & \\
    Definition : You are given a question and some answer options (associated with ""A"", ""B"", ""C"", ""D"", ""E"").  & \\
    You should choose the correct answer based on commonsense knowledge. Avoid answering questions based on associations, & \\ 
    the set of answers are chosen deliberately to capture common sense beyond associations.  & \\
    Do not generate anything else apart from one of the following characters: 'A', 'B, 'C', 'D', 'E' and only give one answer for each question. & \\
     & \\
    Read the examples above and generate task definition for given task type and instance. & \\
    \bottomrule
    \end{tabular}}
\caption{GPT-4 query used for BBH instruction generation.}
\label{table:bbh_query}
\vspace{-3mm}
\end{table*}

\begin{table*}[h!]
    \resizebox{\textwidth}{!}{\begin{tabular}{lc}
    \toprule
    \textbf{Task} & \textbf{Instruction} \\
    \midrule
    Boolean Expressions & In this task, you are provided with a boolean expression composed of boolean values (True or False), \\ 
    & logical operators (and, or, not), and parentheses. Your objective is to evaluate the given boolean \\
    & expression and determine whether the overall expression is True or False. You must strictly adhere \\
    & to the rules of boolean algebra and the precedence of logical operators while evaluating the \\
    & expression. The output should be a single word, either "True" or "False", based on the result of the \\
    & evaluation. No intermediate steps or explanations are required, just the final boolean result.\\
    \midrule
    Causal Judgement & The task of causal judgement involves evaluating a narrative to determine whether a specific factor \\
    & or event caused a particular outcome. You will be provided with a detailed account of events or \\
    & circumstances leading up to an outcome. Based on the information given, you must decide if the factor \\
    & or event in question directly caused the outcome. Your response should be binary, choosing either \\
    & "Yes" if you believe the factor or event caused the outcome, or "No" if you believe it did not. \\
    & You should consider only the information provided in the narrative and not rely on any external\\
    & knowledge or assumptions. The goal is to make a judgement about the direct causality between\\
    & the given factor or event and the outcome.\\
    \midrule
    Date Understanding & In the date understanding task, you are presented with a sentence or sentences that describe how dates\\
    & are formatted or understood in a particular context or region. You are then given a date-related \\
    & question that requires you to apply the information from the input to determine the correct date. The input \\
    & may include a specific date and require you to perform a calculation, such as finding a date a certain \\
    & number of days, weeks, months, or years in the past or future. The output is a selection from \\
    & multiple-choice options, each representing a possible date in a specific format (e.g., MM/DD/YYYY). Your job \\
    & is to choose the option that accurately reflects the answer to the question based on the input information. \\
    & You must use only the information provided in the input and the common knowledge about date calculations \\
    & without assuming any additional facts. Do not generate anything else apart from one of the following\\
    & characters: 'A', 'B, 'C', 'D', 'E', 'F'.\\
    \midrule
    Disambiguation QA & In this task, you are presented with a sentence that contains a pronoun. Your job is to determine the antecedent\\
    & of the pronoun—the specific noun that the pronoun is replacing—or to declare that the antecedent is ambiguous.\\
    & You will be given a sentence and a set of options. Each option will propose a possible antecedent for the\\
    & pronoun in question. You must select the option that correctly identifies the antecedent. If the\\
    & sentence does not provide enough information to determine the antecedent with certainty, you should choose\\
    & the option that indicates the pronoun's antecedent is ambiguous. You are also given 4 answer options\\
    & (associated with "A", "B", "C", "D"), out of which only one is correct. Your output should be the \\
    & letter corresponding to the correct option. \\
        \midrule
    Dyck Languages & In this task, you are presented with a sequence of opening and closing brackets of various types, such as \\
    & parentheses (), square brackets [], and angle brackets <>. Your objective is to complete the sequence by\\
    & adding the appropriate closing brackets in the correct order, ensuring that all brackets are properly matched\\
    & and closed. The input will consist of a partial sequence of brackets, and you must determine the correct\\
    & sequence of closing brackets to complete it. The output should be the minimal sequence of closing\\
    & brackets that, when appended to the input, results in a properly balanced string of brackets with\\
    & all pairs correctly matched. \\
    \midrule
    Formal Fallacies & In this task, you are presented with an argument that consists of premises and a conclusion. Your role is to \\
    & determine whether the argument is deductively valid or invalid based on the explicitly stated premises.\\
    & An argument is considered deductively valid if the conclusion logically follows from the premises, meaning \\
    & that if the premises are true, the conclusion must be true. An argument is deductively invalid if the \\
    & conclusion does not logically follow from the premises, meaning that even if the premises are true, \\
    & the conclusion could still be false. You must choose between two options: "valid" if the argument is \\
    & deductively valid, or "invalid" if the argument is deductively invalid. Do not consider any \\
    & outside knowledge or unstated assumptions; your judgment should be based solely on the information \\
    & provided in the input.\\
    \midrule
    Geometric Shapes & In this task, you are presented with an SVG (Scalable Vector Graphics) path element, which is a string of characters \\
    & that defines the shape of a two-dimensional graphic. The path element is followed by a list of options, each\\
    & representing a different geometric shape. Your job is to identify which geometric shape the given SVG \\
    & path element represents from the provided options. The correct shape must match the structure and \\
    & number of sides as indicated by the SVG path's drawing commands. There is only one correct answer\\
    & from the given list of geometric shapes. \\
        \midrule
    Hyperbaton & The task of hyperbaton involves determining the proper syntactic arrangement of words in a sentence. You are \\
    & presented with multiple sentences, each with a different sequence of adjectives before a noun. Your \\
    & objective is to select the sentence that adheres to the conventional order of adjectives in English. \\
    & The correct order typically follows the sequence: quantity or number, quality or opinion, size, age, shape, \\
    & color, proper adjective (often nationality, other place of origin, or material), and purpose or qualifier. \\
    & Choose the option that places the adjectives in the correct order, resulting in a grammatically coherent \\
    & and standard sentence. Only one option will be the correct sentence with the proper adjective order. \\
    & Answer either 'A' or 'B'. \\
    \bottomrule
    \end{tabular}}
\caption{Generated BBH task instructions using GPT-4[0:8]}
\label{table:bbh_instructions_1}
\vspace{-3mm}
\end{table*}

\begin{table*}[h!]
    \resizebox{\textwidth}{!}{\begin{tabular}{lc}
    \toprule
    \textbf{Task} & \textbf{Instruction} \\
    \midrule
    Logical Deduction & In this task, you are presented with a paragraph that describes a series of some objects with a given attribute,\\
    Three/Five/Seven & such as age, arranged in a specific order. The statements provided are logically consistent and relate to each other \\
    Objects & to form a sequence based on that attribute. Your job is to use logical deduction to determine the relative ordering \\
    & of these objects based on the information given. After analyzing the statements, you must choose the \\
    & correct option from a list that accurately reflects the order of one of the objects in relation to the others. \\
    & There is only one correct answer for each set of statements. \\
    \midrule
    Movie Recommendation & In this task, you are provided with a list of movies and a request to find a movie that is similar to the ones listed.\\
    & The input consists of several movie titles that may share common themes, genres, or elements. You are also \\
    & given a set of options, each representing a different movie. Your task is to select the movie from the options \\
    & that best matches the similarity criteria based on the given list. The output should be the letter corresponding\\
    & to the most similar movie. When determining similarity, consider factors such as plot, genre, themes, \\
    & directorial style, and cultural impact. The response must be one of the provided options, represented by \\
    & a single letter within parentheses. - (A), (B), (C), or (D) \\
    Multistep Arithmetic & In this task, you're presented with a complex arithmetic expression that requires multiple steps to solve. \\
    & Your job is to calculate the value of the entire expression step by step, following the order of operations, \\
    & which is parentheses first, then exponents, multiplication and division from left to right, and finally addition \\
    & and subtraction from left to right. The expression may include negative numbers and a variety of operations. \\
    & Provide the final numerical answer as the output. \\
    \midrule
    Navigate & In this task, you are provided with a set of instructions that describe movements from a starting point. \\
    & The movements can be in different directions and for a specified number of steps. Your task is to analyze\\
    & these instructions to determine if following them would lead you back to the starting point. You must always \\
    & assume that you are facing forward at the beginning and continue to face in the direction of the last\\
    & movement made. You will answer with "Yes" if the instructions lead you back to the starting point,\\
    & or "No" if they do not. Only the options "Yes" or "No" should be provided as the output.\\
    \midrule
    Object Counting & In this task, you are presented with a sentence that lists various objects. Your job is to count the number\\
    & of individual objects mentioned in the sentence. The input will contain the names of the objects and the\\
    & quantities associated with them. You must provide an output that is the total sum of all the objects. Ensure\\
    & that the count is accurate and reflects the information provided in the input. The output should be \\
    & a numerical value representing the total count of all objects listed.\\
    \midrule
    Penguins in a Table & In this task, you are presented with a table of data where the first row contains column headers and each subsequent row\\
    & represents information about a penguin, including its name, age, height in centimeters, and weight in kilograms. \\
    & Your job is to analyze the table and answer a question about the penguins based on the data provided. The question \\
    & will require you to sort or manipulate the data in some way, such as alphabetically sorting the names of the \\
    & penguins. You will be given multiple-choice options to select the correct answer. You are given 5 answer options\\
    & (associated with "A", "B", "C", "D", "E"). The output should be the letter corresponding to the correct\\
    & answer from the provided options. \\
    \midrule
    Reasoning about & You are presented with a scenario that describes the arrangement of various colored objects. Given this scenario, you must \\
    Colored Objects & answer a question that requires logical reasoning to determine the color of a specific object based on its position \\
    & relative to another object. A list of color options, each associated with a letter from (A) to (R), is \\
    & provided. Your task is to select the correct color option that answers the question by indicating the corresponding\\
    & letter. Only one letter should be provided as the answer, and it should accurately reflect the color of the object\\
    & in question as described in the input scenario.\\
        \midrule
    Ruin Names & You are presented with a list of similar-sounding or similarly spelled names based on a given artist or movie name. \\
    & Your task is to identify the option that represents a humorous or intentionally altered version of the original\\
    & name. The input will include the original name and a set of options labeled (A), (B), (C), and (D). Your output\\
    & should be the letter corresponding to the option that is a playful or comical modification of the original\\
    & name. Only select one letter as your answer.\\
    \midrule
    Salient Translation & You are presented with an English translation of a sentence originally in German and a list of potential\\
    Error Detection & types of errors that could be present in the translation. The types of errors include issues with Named Entities,\\
    & Numerical Values, Modifiers or Adjectives, Negation or Antonyms, Facts, and Dropped Content. Your task is to identify\\
    & which type of error is present in the given translation. You must choose from the provided options (A to F) that\\
    & correspond to the types of errors. Your output should be the letter of the option that accurately describes the\\
    & error found in the translation. The goal is to ensure the accuracy of the translation by detecting and\\
    & categorizing the specific error made.\\
    \midrule
    Snarks & In this task, you are provided with a set of statements and asked to identify which one is sarcastic. Sarcasm often \\
    & involves saying the opposite of what is meant, typically for humorous or emphatic effect. Your job is to read\\
    & each option carefully and select the statement that is intended to be taken ironically or in a way that is opposite\\
    & to its literal meaning. The output should be the letter corresponding to the sarcastic statement. Only one\\
    & of the options is considered sarcastic for the purpose of this task. The output will be in the form of A or B,\\
    & corresponding to which option is chosen.\\
    \midrule
    Sports Understanding & In this task, you are presented with a sentence related to sports. Your job is to determine whether the sentence\\
    & is plausible within the context of the sport mentioned. You must answer with "yes" if the sentence could realistically\\
    & occur in the sport, or "no" if it could not. Do not consider any external information or specific events; \\
    & simply assess the plausibility based on general knowledge of the sport. There are only two types of valid \\
    & responses: yes and no.\\
    \bottomrule
    \end{tabular}}
\caption{Generated BBH task instructions using GPT-4[8:21]. Note that Logical Deduction Three/Five/Seven share the same instruction.}
\label{table:bbh_instructions_2}
\vspace{-3mm}
\end{table*}

\begin{table*}[h!]
    \resizebox{\textwidth}{!}{\begin{tabular}{lc}
    \toprule
    \textbf{Task} & \textbf{Instruction} \\
    \midrule
    Temporal Sequences & You are given a scenario that includes a series of events with associated times, and a question that asks you\\
    & to determine the possible time range for a particular event within that sequence. Additionally, you are provided\\
    & with a set of time range options labeled "(A)", "(B)", "(C)", and "(D)". Your task is to select the \\
    & correct time range in which the event could have occurred based on the information provided in the scenario. You\\
    & should only answer with the choice letter that corresponds to the correct time range, without providing additional\\
    & explanation or the full text of the option.\\
    \midrule
    Tracking Shuffled & In this task, you are presented with a scenario involving a group of individuals who are initially paired with partners.\\
    Objects (3)/(5)/(7) & As the scenario unfolds, these individuals swap partners multiple times. Your task is to track the sequence of swaps\\
    & and determine the final partner of a specified individual from a list of options labeled "A" through "E". You should\\
    & analyze the given sequence of events carefully and provide the correct answer based on the final arrangement. Only\\
    & one option is the correct answer, and you should generate the corresponding letter (A, B, C, D, or E) without\\
    & any additional information or explanation.\\
    \midrule
    Web of Lies & In the "web of lies" task, you are presented with a series of statements involving multiple individuals, each making a claim\\
    & about another individual's truthfulness or dishonesty. Your objective is to determine the truthfulness of the \\
    & final individual mentioned based on the information provided in the input statements. The output should be "Yes" \\
    & if you conclude that the final individual is telling the truth, and "No" if you conclude they are lying. To solve\\
    & this task, you must analyze the chain of claims to infer the credibility of each individual, leading to a conclusion\\
    & about the final individual's honesty. Only "Yes" or "No" are valid responses.\\
    \midrule
    Word Sorting & In this task, you are provided with a list of words. Your task is to rearrange these words in alphabetical order,\\
    & starting with the word that comes first in the alphabet and ending with the word that comes last. You should \\
    & generate the sorted list in a single line, with each word separated by a space. Do not add any words that are\\
    & not included in the original list, and do not omit any words from the original list. The output should consist \\
    & solely of the given words, sorted alphabetically.\\
    \bottomrule
    \end{tabular}}
\caption{Generated BBH task instructions using GPT-4[21:27]. Note that Tracking Shuffled Objects (3)/(5)/(7) share the same instruction except for the label space.}
\label{table:bbh_instructions_3}
\vspace{-3mm}
\end{table*}

\end{document}